\documentclass[twocolumn,journal]{IEEEtran}
\usepackage[square, comma, sort&compress, numbers]{natbib}
\usepackage{epsfig}
\usepackage{epstopdf}
\usepackage{graphicx}
\usepackage{citesort}
\usepackage{amssymb}
\usepackage{amsmath}
\usepackage{color}
\usepackage{url}
\usepackage{array}
\usepackage{multirow}
\usepackage[table]{xcolor}
\usepackage{bm}
\usepackage{tablefootnote}
\usepackage{threeparttable}
\usepackage{color}
\usepackage{colortbl}
\usepackage{mdwmath,mdwtab,amsmath}
\usepackage{autobreak}
\usepackage{amsmath,amssymb}
\usepackage{array}
\usepackage{multirow}
\usepackage{tabularx}
\usepackage{makecell}
\usepackage{graphicx}
\usepackage{threeparttable}
\usepackage{tablefootnote}
\usepackage[OT1]{fontenc}
\usepackage{float}
\newcolumntype{Y}{>{\centering\arraybackslash}X}
\usepackage[ruled,linesnumbered]{algorithm2e}
\usepackage{tabularx}
\usepackage[hang,flushmargin]{footmisc}
\usepackage{ragged2e}
\usepackage{fancyhdr}
\usepackage{xcolor}
\usepackage{bbding}

\usepackage{xpatch}
\makeatletter
\ExplSyntaxOn
\cs_new:Npn \bibColoredItems #1#2
{
	\clist_map_inline:nn {#2} { \cs_new:cpn {bib@colored@##1} {#1} } 
}
\ExplSyntaxOff
\newcommand\bib@setcolor[1]{%
	\ifcsname bib@colored@#1\endcsname
	\expandafter\color\expandafter{\csname bib@colored@#1\endcsname}
	\else
	\normalcolor
	\fi
}
\xpatchcmd\@bibitem
{\item}
{\bib@setcolor{#1}\item}
{}{\fail}

\xpatchcmd\@lbibitem
{\item}
{\bib@setcolor{#2}\item}
{}{\fail}
\makeatother

\newlength{\figurewidth}
\newlength{\smallfigurewidth}

\usepackage{tikz,xcolor,hyperref}
\definecolor{lime}{HTML}{A6CE39}
\DeclareRobustCommand{\orcidicon}{
	\begin{tikzpicture}
		\draw[lime, fill=lime] (0,0)
		circle[radius=0.16]
		node[white]{{\fontfamily{qag}\selectfont \tiny \.{I}D}}; %node[white]{{\fontfamily{qag}\selectfont \tiny $\dot{\mathsf I}$D}};
	\end{tikzpicture}
	\hspace{-2mm}
}
\foreach \x in {A, ..., Z}{%
	\expandafter\xdef\csname orcid\x\endcsname{\noexpand\href{https://orcid.org/\csname orcidauthor\x\endcsname}{\noexpand\orcidicon}}
}

\hypersetup{hidelinks}

\begin{document}
	\title
	{A General Gaussian Heatmap Label Assignment for Arbitrary-Oriented Object Detection}
	%%%%%%%%%%%%%%%%%%%%%%%%%%%%%%%%%%%%%%%
	%% Two Column
	%%%%%%%%%%%%%%%%%%%%%%%%%%%%%%%%%%%%%%%
	\author{%
		%}
		\vspace{0.5em}
		Zhanchao Huang, %\hspace{-1.5mm}\orcidA{}
		Wei Li,~\IEEEmembership{Senior Member,~IEEE},
		Xiang-Gen Xia,~\IEEEmembership{Fellow,~IEEE},  \\
		and
		Ran Tao,~\IEEEmembership{Senior Member,~IEEE}
		\thanks{%
			This work was supported by National Key R\&D Program of China under Grant No.2021YFB3900502, the National Natural Science Foundation of China under Grant 61922013 and U1833203, and by the Beijing Natural Science Foundation under Grant L191004 and JQ20021. (Corresponding Author: Wei Li; e-mail: liwei089@ieee.org)}
		\thanks{%
			Zhanchao Huang, Wei Li, Ran Tao are with the School of Information and Electronics, Beijing Institute of Technology, and Beijing Key Lab of Fractional Signals and Systems, 100081 Beijing, China. (e-mail: zhanchao.h@outlook.com; liwei089@ieee.org; rantao@bit.edu.cn).}
		\thanks{%
			Xiang-Gen Xia is with the Department of Electrical and Computer Engineering, University of Delaware, Newark, DE 19716, USA (e-mail: xxia@ee.udel.edu).}
	}
	%%%%%%%%%%%%%%%%%%%%%%%%%%%%%%%%%%%%%%%
	%% Two Column
	%%%%%%%%%%%%%%%%%%%%%%%%%%%%%%%%%%%%%%%
\maketitle
\thispagestyle{fancy}
\renewcommand{\headrulewidth}{0pt} %改为0pt即可去掉页眉下面的横线
\pagestyle{fancy}

\begin{abstract}
Recently, many arbitrary-oriented object detection (AOOD) methods have been proposed and attracted widespread attention in many fields. However, most of them are based on anchor-boxes or standard Gaussian heatmaps. Such label assignment strategy may not only fail to reflect the shape and direction characteristics of arbitrary-oriented objects, but also have high parameter-tuning efforts. In this paper, a novel AOOD method called General Gaussian Heatmap Label Assignment (GGHL) is proposed. Specifically, an anchor-free object-adaptation label assignment (OLA) strategy is presented to define the positive candidates based on two-dimensional (2-D) oriented Gaussian heatmaps, which reflect the shape and direction features of arbitrary-oriented objects. Based on OLA, an oriented-bounding-box (OBB) representation component (ORC) is developed to indicate OBBs and adjust the Gaussian center prior weights to fit the characteristics of different objects adaptively through neural network learning. Moreover, a joint-optimization loss (JOL) with area normalization and dynamic confidence weighting is designed to refine the misalign optimal results of different subtasks. Extensive experiments on public datasets demonstrate that the proposed GGHL improves the AOOD performance with low parameter-tuning and time costs. Furthermore, it is generally applicable to most AOOD methods to improve their performance including lightweight models on embedded platforms.

\end{abstract}


\section{Introduction}
In the past few years, the continued innovation of convolutional neural network (CNN) based object detection (OD) methods has emerged \cite{renFasterRCNNRealTime2017a,redmonYouOnlyLook2016a,tian2019fcos,zhang2021dense}. As one of the more specialized tasks of OD, the AOOD task also follows the trend and develops rapidly. It detects objects more accurately through bounding boxes with directions in scenes of remote sensing \cite{xiaDOTALargeScaleDataset2018,chengLearningRotationInvariantFisher2019, li2020object}, retail \cite{pan2020dynamic}, text \cite{liao2018textboxes}, etc.

Along with the intensive studies, the CNN structure of AOOD models has become more and more complicated to make the distribution of extracted features approximate to the distribution of ground truth. However, it is not the only way to improve the detection performance through extracting features using a CNN structure \cite{zhang2020bridging} as we shall see below.  As shown in step 2 of Fig.~\ref{fig:1}, more than one location in a feature map that is used to detect the object. In this regard, most CNN-based OD methods \cite{renFasterRCNNRealTime2017a,redmonYouOnlyLook2016a,tian2019fcos} assign a label to many candidate locations as ground truth during the CNN training to improve the robustness, which is called label assignment \cite{zhang2020bridging}. From a more macro perspective, the CNN training is essentially a process of learning the one-to-many mapping from the predictions of many candidate locations to a labeled object. Different one-to-many label assignment strategies directly affect the detection performance by generating different ground truths (called sample spaces) for training. Therefore, to improve the detection performance, one way is to use a more complex CNN, i.e., a more complex approximation function. The other way is to design a label assignment strategy for constructing a better sample space that is more in line with the characteristics of the object's shape and direction. The latter is as important as the former in object detection tasks.
\begin{figure}[tbp]
	\vspace{-1em}
	\centering
	\epsfig{width=0.42\textwidth,file=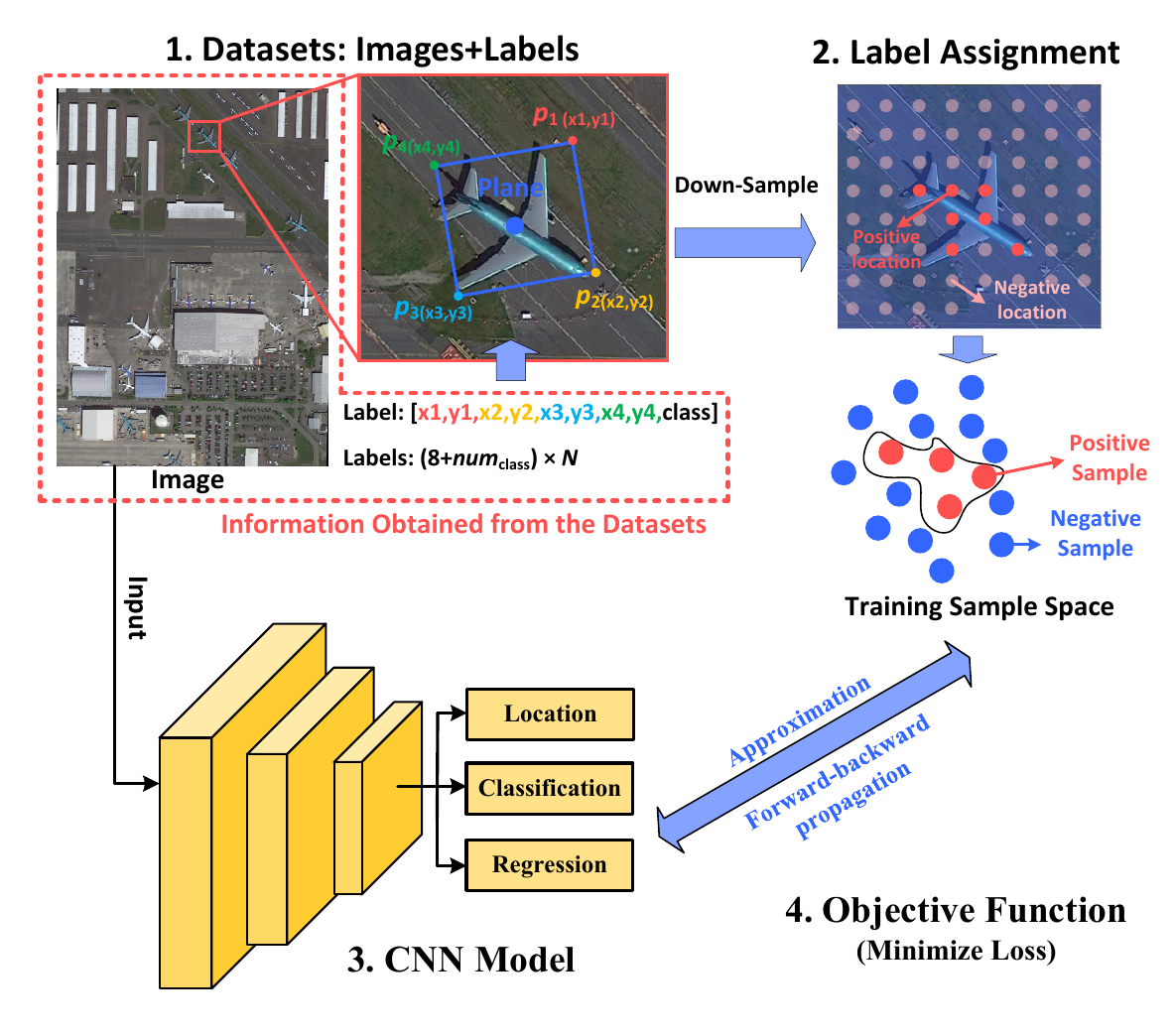}
	\vspace{-1.5em}
	\caption{The primary process of training a CNN-based object detection model that adopts the dense detection paradigm.}\label{fig:1}
	\vspace{-1em}
\end{figure}

\begin{figure}[tbp]
	%\vspace{-1em}% 调整间距
	\centering
	\epsfig{width=0.45\textwidth,file=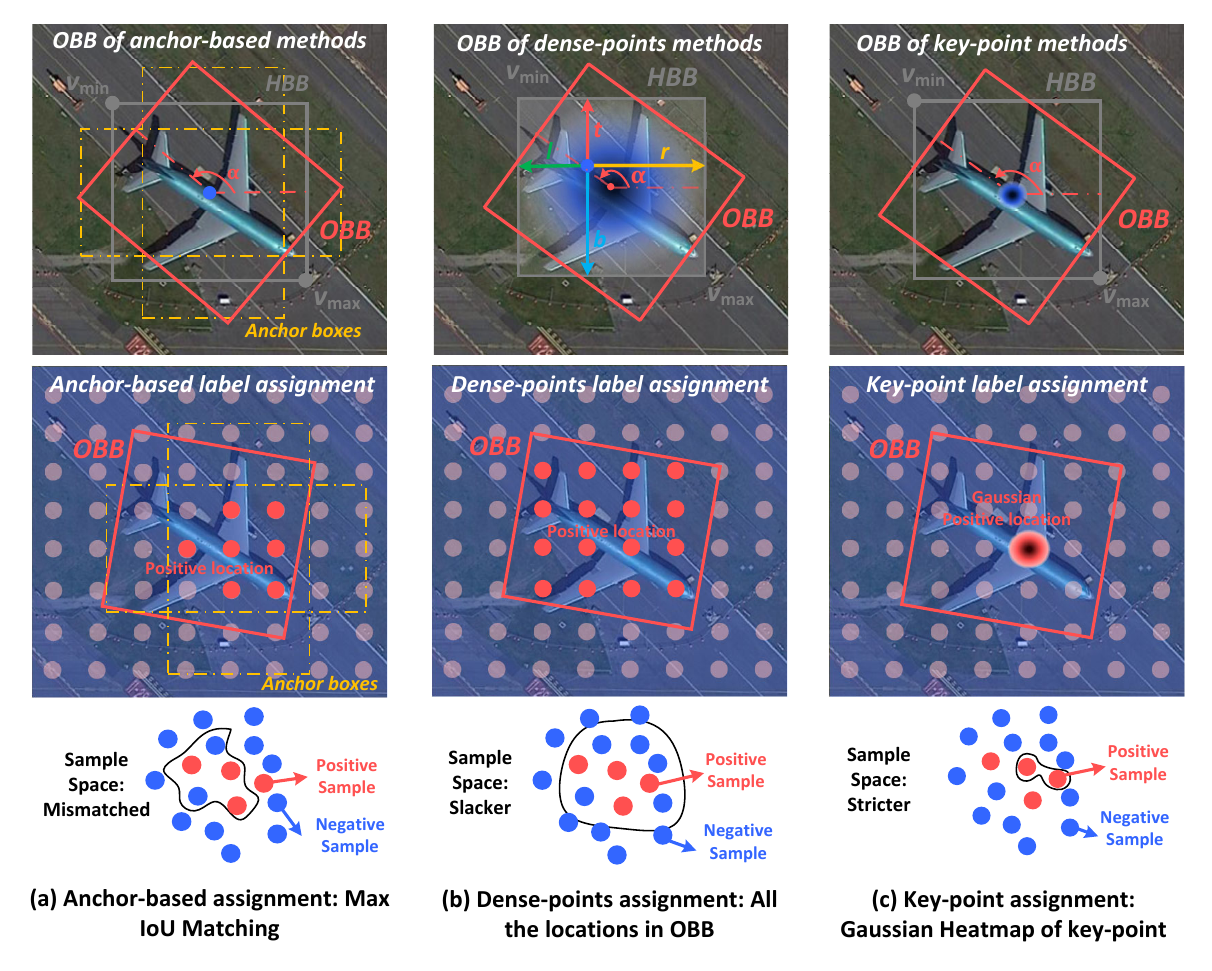}
	\vspace{-1.5em}
	\caption{Schematic diagram of mainstream label assignment strategies and their sample spaces. (a) The anchor-based strategy. (b) The dense-points strategy. (c) The key-point strategy.}\label{fig:2}
	\vspace{-1em}
\end{figure}

Most of the AOOD methods, such as SCRDet \cite{yangSCRDetMoreRobust2019}, LO-Det \cite{huang2021lo}, DAL\cite{ming2020dynamic}, CenterMap\cite{wang2020learning}, DCL \cite{yang2021dense}, Oriented R-CNN \cite{xie2021oriented}, etc., use the anchor-based label assignment strategy, as shown in Fig.~\ref{fig:2} (a). However, this strategy may lead to the mismatch of positive and negative (P\&N) locations when the default anchor boxes cannot cover a specific shape \cite{zhang2020bridging}, especially in the complex scene. Besides, the anchor-based strategy requires many dataset-dependent hyperparameters \cite{zhang2021learning}, which costs a lot of efforts for tuning when the dataset is changed \cite{zhu2020autoassign}. Regarding the above issues, anchor-free methods like FCOS \cite{tian2019fcos} and CenterNet \cite{zhou2019objects} redefine P\&N locations \cite{zhang2020bridging} and get rid of the dependence on anchors' shape. Among them, dense-points methods, such as FCOS \cite{tian2019fcos}, IENet \cite{lin2019ienet}, AOPG \cite{cheng2021anchor}, etc., relax the sample space constraints, as shown in Fig.~\ref{fig:2} (b), which may cause some negative locations to be misallocated as positive locations. While the key-point methods like CenterNet \cite{zhou2019objects}, BBAVectors \cite{yi2020oriented}, O$^2$-DNet \cite{wei2020oriented}, etc., use a stricter positive location assignment strategy, as shown in Fig.~\ref{fig:2} (c), which relies on higher resolution feature maps and makes the number of P\&N locations more unbalanced. Furthermore, the above label assignment strategies do not fully consider the characteristics of the object’s shape and direction when defining P\&N locations. Therefore, the expected label assignment strategy should construct sample space without anchor boxes and define P\&N locations that are more in line with the characteristics of objects in the AOOD task. 

An expected training sample space also requires a proper objective function to guide the model to learn higher quality features, as shown in Fig.~\ref{fig:1}. Whether it is OD or AOOD task, the mainstream objective function paradigm is to optimize classification and OBB regression tasks independently and minimize the sum of each loss as the optimization goal \cite{cao2020prime}. However, there may be one of the two situations: an accurately localized object has a lower classification score, or an accurately classified object has a lower OBB regression score. Therefore, as analyzed by PISA \cite{cao2020prime}, Free-Anchor \cite{zhang2021learning}, AutoAssign \cite{zhu2020autoassign}, etc., jointly optimizing different subtasks is a more reasonable goal. Furthermore, it should be considered that different objects may have different numbers of positive locations, and different locations contribute unequally to the loss function.

In summary, mismatched, overly slack, or unduly strict label assignment strategies make it difficult for constructed training sample space to adapt to object characteristics and may have many hyperparameters. Moreover, the inconsistent loss function for different subtasks makes it more challenging to learn the optimal parameters of CNN in the sample space. Therefore, for a CNN-based AOOD method, compared with developing a dazzling network structure, it is even more important to construct an object-adaptation label assignment strategy and design a goal-consistent loss function. In this regard, a novel and practical AOOD method with higher performance and fewer hyperparameters called General Gaussian Heatmap Label Assignment (GGHL) is proposed. The contributions of this work are summarized as follows:

1) An object-adaptation label assignment (OLA) strategy without any prior anchor boxes is proposed based on two-dimensional (2-D) oriented Gaussian heatmaps. It simplifies the P\&N location definition and makes the distribution of positive locations more flexible to fit the object's size and direction.

2) An oriented-bounding-box representation component (ORC) based on the distances from the positive point to OBB vertexes is developed, which indicates any OBBs without anchor boxes. Furthermore, an object-adaptive weight adjustment mechanism (OWAM) is designed to adaptively adjust the Gaussian center prior weights of different locations and used to weight the loss of different P\&N locations.

3) A joint-optimization loss (JOL) with area normalization and dynamic weighting is proposed. It refines the misaligned optimization goals between positive and negative locations, OBB regression and classification tasks by jointly optimizing their likelihood function (LF). Besides, it balances the model's learning preferences for objects of different categories with different sizes at different locations.

The remainder of this paper is organized as follows. Section II reviews and analyzes the related works. Section III presents a detailed description of the proposed GGHL. In Section IV, extensive experiments are conducted, and the results are discussed. Finally, conclusions are summarized in Section V.

%\cite{howard2017mobilenets,sandler2018mobilenetv2,howard2019searching}

\section{Related Works}
\subsection{Arbitrary-Oriented Object Detection}
Benefiting from the open-source AOOD datasets annotated with OBBs in the scenes like remote sensing \cite{xiaDOTALargeScaleDataset2018}, the prediction of the OD model has become more refined, which helps to accurately locate the object in the image and reflect its shape and direction. In the AOOD task, whether the two-stage methods \cite{yangSCRDetMoreRobust2019,ding2019learning,xu2020gliding} or the one-stage methods \cite{pan2020dynamic,huang2021lo,ming2020dynamic}, most of them adopt the anchor-box-based framework due to its mature application in various OD tasks. However, since oriented anchors are more prone to mismatch problems and have more hyperparameters than horizontal anchors, many works have dealt with them. For example, Ding et al. \cite{ding2019learning} transformed ROI to rotated ROI for avoiding a large number of oriented anchors in the two-stage detector. Xu et al. \cite{xu2020gliding} proposed a gliding vertex method to represent OBBs, the model of which is based on horizontal anchors without setting oriented anchors with multiple angles. DAL \cite{ming2020dynamic} analyzed and proposed a dynamic matching and assignment strategy. Oriented R-CNN \cite{xie2021oriented} proposed an oriented RPN to directly generate oriented proposals in a nearly cost-free manner and employed the midpoint offsets to represent OBBs based on Gliding Vertex \cite{xu2020gliding}. To remove anchor boxes, BBAVectors \cite{yi2020oriented}, DRN \cite{pan2020dynamic}, O$^2$-DNet \cite{wei2020oriented}, etc., employed the anchor-free framework and designed new OBB representation components. AOPG \cite{cheng2021anchor} abandoned the horizontal boxes-related operations and generated oriented boxes by Coarse Location Module in an anchor-free manner. However, these anchor-free AOOD methods do not consider the characteristics of the object's shape and direction while just borrowing label assignment strategies from other OD tasks. In addition, few other methods like CSL \cite{yang2020arbitrary} predict oriented objects through angle classification. Fig.~\ref{fig:2} summarizes the mainstream label assignment and OBB representation strategies of the existing AOOD methods.

\vspace{-0.5em}
\subsection{Label Assignment Strategy}
The label assignment is a core issue that a CNN model based on the dense-positive detection paradigm needs to consider. Faster R-CNN \cite{renFasterRCNNRealTime2017a} introduces the anchor-based label assignment strategy to explicitly enumerate the prior information of different scales and aspect ratios. This strategy introduces many hyperparameters that depend on the datasets \cite{zhang2021learning}. It means that one needs to spend a lot of efforts adjusting the hyperparameters when the dataset is changed \cite{zhu2020autoassign}. Moreover, these easily overlooked recessive costs cannot be reduced by a lightweight CNN model \cite{huang2021lo}. To solve the problem that the anchor-based strategy relies on many hyperparameters and may have mismatches, some OD methods, such as FCOS \cite{tian2019fcos} and CenterNet \cite{zhou2019objects}, designed different anchor-free assignment strategies, as shown in Fig.~\ref{fig:2}. ATSS \cite{zhang2020bridging} analyzed and suggested that the gap between the anchor-box-based strategy and the anchor-free strategy lies in the definition of P\&N locations. Borrowing from the learning-to-match strategy of FreeAnchor \cite{zhang2021learning}, AutoAssign \cite{zhu2020autoassign} further let the model learn to define P\&N locationcs and assign labels automatically. However, the existing label assignment strategies do not fully consider the characteristics of the object location, shape, and direction when defining P\&N locations. Therefore, how to design a more appropriate label allocation strategy for oriented objects remains to be explored.

\vspace{-0.5em}
\subsection{Loss Function for AOOD}
Most existing AOOD methods still follow the classic OD loss paradigm that optimizes the OBB regression and object classification tasks separately \cite{chen2020piou,ming2021optimization,yang2021rethinking}. The difference between them and ordinary OD loss is an additional loss related to the direction of OBBs. For instance, PIoU \cite{chen2020piou} calculated the approximate IoU of the OBB and ground truth through pixel counting; RIL \cite{ming2021optimization} used the Hungarian algorithm to determine the optimal matching; GWD \cite{yang2021rethinking} represented the OBB regression loss by the distance of the Gaussian distributions. Furthermore, KLD \cite{yang2021learning} used the Kullback-Leibler divergence between the Gaussian distributions as the regression loss, which dynamically adjusted the parameter gradients according to the characteristics of the object. While DCL \cite{yang2021dense} further optimizes the accuracy and efficiency of angle classification loss based on CSL \cite{yang2020arbitrary}. Although these methods are effective, they did not consider the inconsistency of OBB regression and object classification optimization goals and relied on a large number of anchor boxes. In ordinary OD tasks, although PISA \cite{cao2020prime}, FreeAnchor \cite{zhang2021learning}, AutoAssign \cite{zhu2020autoassign}, etc., analyzed this problem, they did not consider the direction and shape of OBBs and were not used in the AOOD task. Moreover, they did not notice that the contributions to the loss function of different objects and different locations are different, which needs to be studied.

\section{Proposed GGHL Framework}
The framework of the proposed GGHL is shown in Fig. 3, which is mainly composed of three parts: the proposed label assignment strategy OLA, the CNN model with developed ORC, and the designed objective function JOL. First, each label is assigned one-to-many to the Gaussian candidate locations in the feature maps through the proposed OLA strategy. Second, a CNN model is constructed to extract features from the input images. Then, the proposed ORC encodes these features to predict the OBB and category at each positive location. Furthermore, the Gaussian prior weight of each positive candidate location is adjusted by the designed CNN-learnable OWAM to fit the object’s shape adaptively. Third, the joint-optimization loss between the ground truth of the constructed training sample space and the prediction of the CNN model is calculated. Finally, the CNN model is trained until the loss converges to obtain the optimal parameters.
\begin{figure}[tbp]
	\vspace{-0.5em}
	\centering
	\epsfig{width=0.45\textwidth,file=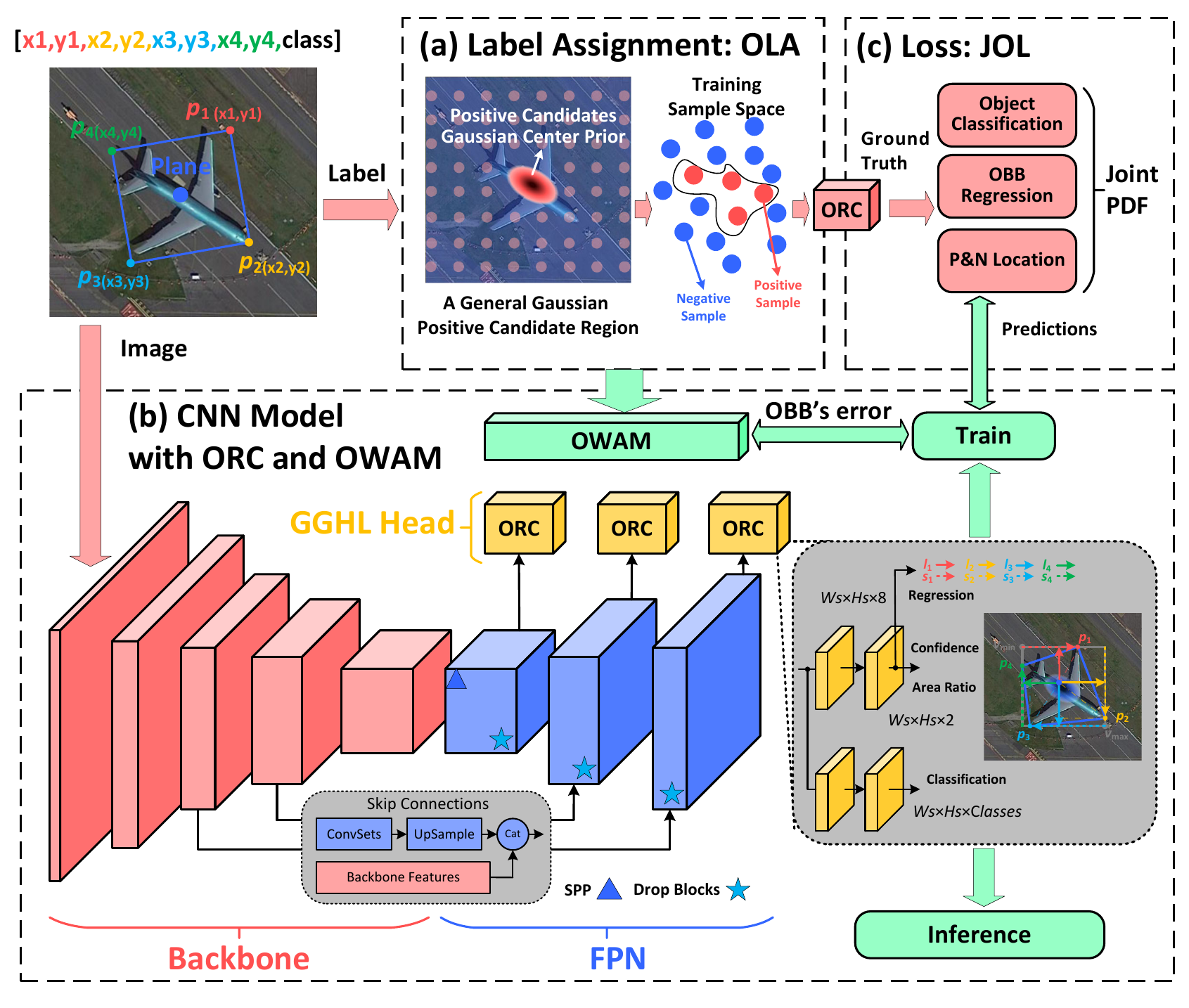}
	\vspace{-1em}
	\caption{The GGHL framework comprises (a) the proposed OAL strategy, (b) the CNN model with developed ORC and OWAM, (c) the designed objective function JOL.}\label{fig:3}
	\vspace{-0.5em}
\end{figure} 

\vspace{-0.5em}
\subsection{OLA Strategy}

\begin{figure}[tbp]
	\vspace{-1em}
	\centering
	\epsfig{width=0.48\textwidth,file=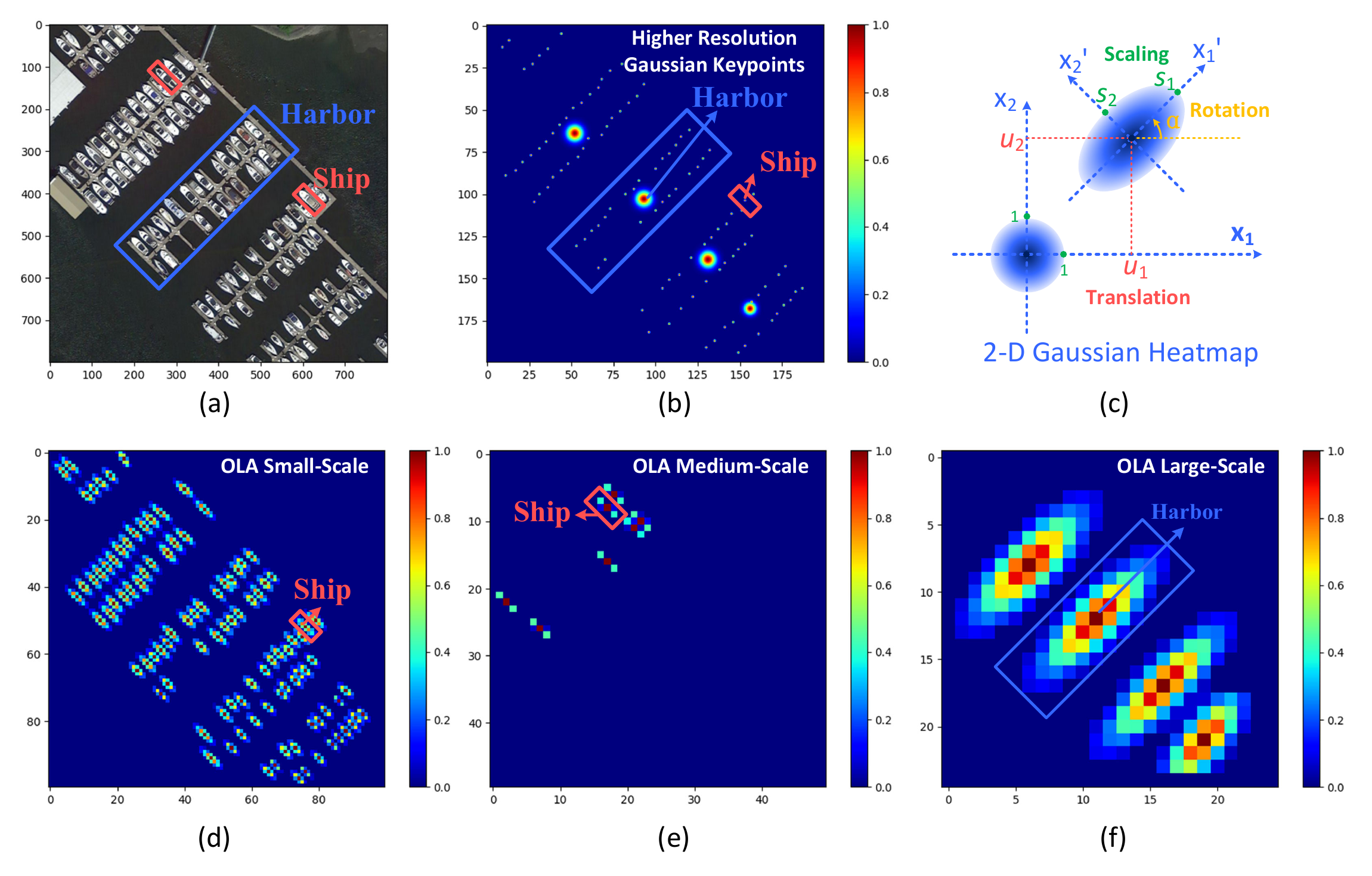}
	\vspace{-1em}
	\caption{The principle of the proposed object-adaptation label assignment (OLA) strategy. (a) The original image. (b) The higher-resolution heatmaps (down-sampling stride=4) generated by the Gaussian key-points strategy of CenterNet \cite{zhou2019objects}, BBAVectors \cite{yi2020oriented}, DRN \cite{pan2020dynamic}, etc. (c) The principle of generating 2-D Gaussian heatmaps. (d) (e) (f) represent the multiscale (down-sampling stride=8, 16, 32) Gaussian positive candidates generated by the proposed OLA strategy. The color bars represent Gaussian probability.}
	\label{fig:4}
	\vspace{-1em}
\end{figure}

In the previous works, although anchor-based methods, e.g., CenterMap \cite{wang2020learning}, introduced "Gaussian-like" or "Centerness-like" \cite{tian2019fcos} weighting mechanisms for positive candidates, they are still essentially based on maximum IoU matching to define P\&N samples. As mentioned before, such matching strategies suffer from mismatch risks especially in dense object scenarios and they rely on a large number of hyperparameters. Methods like GWD \cite{yang2021rethinking} mainly employ 2-D Gaussians for loss calculation, and its label assignment is still based on anchor boxes. CenterNet \cite{zhou2019objects}, BBAvectors \cite{yi2020oriented}, DRN \cite{pan2020dynamic}, etc., also use the 2-D Gaussian distribution to define positive candidate locations, as shown in Fig.~\ref{fig:4} (b). However, their Gaussian heatmap of each object is a circle (the standard Gaussian distribution), which may not well reflect the shape and direction of an object. Besides, they only take Gaussian peak points as positive locations, which need to detect objects on higher resolution feature maps (stride=4) with higher computational complexity and more unbalanced P\&N locations. In contrast, the proposed OLA uses an oriented elliptical Gaussian region to represent an object's positive candidate set intuitively. Furthermore, the objects are assigned to lower-resolution feature maps with different scales (stride=8, 16, 32) according to their sizes, as shown in Fig.~\ref{fig:4} (d) (e) (f), which has lower computational complexity and is compatible with the mainstream Backbone-FPN \cite{linFeaturePyramidNetworks2017a,redmonYOLOv3IncrementalImprovement2018} pipeline in OD tasks. 

Different from the existing methods, the proposed OLA strategy directly uses the 2-D Gaussian for label assignment to make the assigned candidates more in line with objects’ shapes and directions, and alleviates the mismatch problem of anchor-based methods in dense instance scenarios. More specifically, the proposed OLA more fully discusses the relationship between 2-D Gaussian and geometric transformations in oriented object label assignment from a theoretical perspective. Based on this, the technical details to be considered for 2-D Gaussian label assignment are explained, including multi-scale assignment, overlap problem in the assignment, discussion of Gaussian radius, etc.

\textbf{1) First, a general 2D Gaussian distribution is used to represent the positive candidate area with rotation and scaling}, and the locations of the entire Gaussian region are regarded as positive locations and given different weights according to the Gaussian density function, compared to just taking the point at Gaussian peak as the positive locations.

Specifically, the Gaussian probability density function (PDF) is represented as
\begin{equation}
	f\left(\boldsymbol X \right) = \frac{1}{{\sqrt {2\pi \boldsymbol C} }} \times {e^{ - \frac{1}{2}{{\left( {\boldsymbol X - \boldsymbol u} \right)}^T}{\boldsymbol C^{ - 1}}\left( {\boldsymbol X - \boldsymbol u} \right)}},
\label{eq:1}
\end{equation}
where  $\boldsymbol X = {\left[ {x,y} \right]^T} \sim N\left( {\boldsymbol \mu ,\boldsymbol C} \right)$ contains two random variables in the two dimensions. $\boldsymbol \mu  \in {R^2}$ represents the mean vector, and the non-negative semi-definite real matrix $\boldsymbol C \in {R^{2 \times 2}}$ represents the covariance matrix of the two variables. The real symmetric matrix $\boldsymbol C$ is orthogonally diagonalized and decomposed into
\begin{equation}
	\boldsymbol C = \boldsymbol A{\boldsymbol A^T} = \boldsymbol Q \boldsymbol \Lambda {\boldsymbol Q^T} = \left( {\boldsymbol Q{\boldsymbol \Lambda ^{1/2}}} \right){\left( {\boldsymbol Q{\boldsymbol \Lambda ^{1/2}}} \right)^T}.
	\label{eq:2}
\end{equation}
Thus,
\begin{equation}
	\begin{array}{l}
	{\left( {\boldsymbol X - \boldsymbol \mu } \right)^T}{\boldsymbol C^{ - 1}}\left( {\boldsymbol X - \boldsymbol \mu } \right) = \\
	{\left[ {{{\left( {\boldsymbol Q{\boldsymbol \Lambda ^{1/2}}} \right)}^T}\left( {\boldsymbol X - \boldsymbol \mu } \right)} \right]^T}\left[ {{{\left( {\boldsymbol Q{\boldsymbol \Lambda ^{1/2}}} \right)}^T}\left( {\boldsymbol X - \boldsymbol \mu } \right)} \right],
\end{array}
	\label{eq:3}
\end{equation}
where $\boldsymbol Q$ is a real orthogonal matrix, and $\boldsymbol \Lambda$ is a diagonal matrix composed of the eigenvalues of descending order. The Gaussian probability density function is transformed into
\begin{equation}
	\begin{array}{l}
	\! f\left(\boldsymbol X \right) \!= \! \frac{1}{{\sqrt {2\pi \boldsymbol Q\boldsymbol \Lambda {\boldsymbol Q^T}} }} \! \times {\! e^{ \! - \! \frac{1}{2}{{\! \left[ {{{\left( {\boldsymbol Q{\boldsymbol \Lambda \! ^{\frac{1}{2}}}} \! \right)}\! ^T}\! \left( {\! \boldsymbol X \! - \! \boldsymbol \mu } \! \right)} \! \right]}\! ^T} \! \left[ {{{\! \left( {\! \boldsymbol Q{\boldsymbol \Lambda \! ^{\frac{1}{2}}}} \right)}\! ^T}\! \left( {\! \boldsymbol X \! - \! \boldsymbol \mu } \right)} \! \right]}}.
	\end{array}	
\label{eq:4}
\end{equation}

From the perspective of geometric transformation, the mean vector $\boldsymbol \mu  = {\left[ {{\mu _1},{\mu _2}} \right]^T}$ controls the spatial translation. The real orthogonal matrix $\boldsymbol Q$ is a rotation in this case:
\begin{equation}
	\boldsymbol Q = \left[ {\begin{array}{*{20}{c}}
		{\cos \alpha }&{ - \sin \alpha }\\
		{\sin \alpha }&{\cos \alpha }
\end{array}} \right],
\label{eq:5}
\end{equation}
where $\alpha$ denotes the angle of rotation. Because $-\boldsymbol Q$ and $\boldsymbol Q$ are the same in this case, $\alpha  \in \left[ {0,\pi } \right)$. The diagonal matrix $\boldsymbol \Lambda$ composed of eigenvalues represents the scaling, that is
\begin{equation}\boldsymbol \Lambda  = \boldsymbol S{\boldsymbol S^T} = \left[ {\begin{array}{*{20}{c}}
		{{\lambda _1}}&{}\\
		{}&{{\lambda _2}}
\end{array}} \right] = \left[ {\begin{array}{*{20}{c}}
		{s_1^2}&{}\\
		{}&{s_2^2}
\end{array}} \right],
\label{eq:6}
\end{equation}
where the eigenvalues $\lambda_1$ and $\lambda_2$ represent the square of the semi-major axis $s_1$ and the square of the semi-minor axis $s_2$ of the ellipse, respectively. Finally, the distribution becomes the standard Gaussian distribution of ${\left[ {0,0} \right]^T}$ mean vector and ${\boldsymbol{I}_{2 \times 2}}$ covariance matrix, where ${\boldsymbol{I}_{2 \times 2}}$ is the 2×2 identity matrix. 

\begin{algorithm}[!t]
	\label{alg:1}
	\small%\footnotesize
	\caption{Generate the Gaussian Candidate Region}%算法名字	
	\LinesNumbered %要求显示行号
	\KwIn{Labels, each of which contains four vertices ($\left( {{x_1},{y_1}} \right)$, $\left( {{x_2},{y_2}} \right)$, $\left( {{x_3},{y_3}} \right)$, $\left( {{x_4},{y_4}} \right)$) of the OBB:  represent an OBB); number of labels ${N_l}$}%输入参数
	\KwOut{General Gaussian heatmap $\boldsymbol F$}%输出
	\For{1 to ${N_l}$}{
		Pre-process the label to get $\boldsymbol \mu$, $\boldsymbol Q$, $\boldsymbol \Lambda$ \;
		Calculate the threshold $thr = f\left( {{x_b},{y_b}} \right)$ at the end 
		point $\left( {{x_b},{y_b}} \right)$ of the semi-axis according to $\boldsymbol \Lambda$ and Eq.~\ref{eq:4}, which is explained in Section III-A-2) \;
		\For{$\min ({x_1},{x_2},{x_3},{x_4})$ to $\max ({x_1},{x_2},{x_3},{x_4})$}{
			\For{$\min ({y_1},{y_2},{y_3},{y_4})$ to $\max ({y_1},{y_2},{y_3},{y_4})$}{
				Calculate $f\left( {x,y} \right)$ according to Eq.~\ref{eq:4} \;
				\If{$f\left( {x,y} \right) < thr$}{
					${F_{x,y}} = 0$\;
				}
				\If{$f\left( {x,y} \right) > {F_{x,y}}$}{
					${F_{x,y}} = f\left( {x,y} \right)$
					Assign other parameters of the label (see Section III-B)\;
				}
			}	
		}
		Normalize $f\left( {x,y} \right)$ in each Gaussian region.
	}
\end{algorithm}

In summary, as shown in Fig.~\ref{fig:4} (c), probability density of any 2-D Gaussian distribution $f\left(\boldsymbol X \right)$ is obtained by a linear transformation from a standard 2-D Gaussian distribution (two random variables are independent Gaussian random variables with normal distribution). According to Eq.~\ref{eq:4}, a P\&N location distribution map $\boldsymbol F$ is generated through Algorithm~\ref{alg:1}. Define the element at $\left( {x,y} \right)$ of $\boldsymbol F$ as $F_{x,y}$; and define $f\left( {x,y} \right) \in \left[ {0,1} \right]$ as the Gaussian value at $F_{x,y}$ calculated by Eq.~\ref{eq:4}, which is normalized in each generated Gaussian region respectively. If $f\left( {x,y} \right) = 0$, this location is defined as negative (background), ${F_{x,y}} = 0$. If $f\left( {x,y} \right) > 0$, this location is defined as positive (foreground), ${F_{x,y}} = f\left( {x,y} \right)$, and the value of $f\left( {x,y} \right)$ represents the weight of this location in the Gaussian region it belongs to. 

\textbf{2) Second, the problem of possible overlap of Gaussian regions needs to be considered in the assignment process.} Unlike FCOS \cite{tian2019fcos} or CenterMap \cite{wang2020learning}, which assign overlapping region labels to instances with smaller areas, the proposed OLA allows more flexibility to assign labels for each candidate location. Specifically, if a location is contained in different Gaussian regions, it is assigned to the region that has the largest $f\left( {x,y} \right)$. This location is selected as the candidate to predict the object belongs to this Gaussian region. Moreover, the calculation of weights using Gaussian PDFs does not suffer from interpolation approximation problems encountered during the rotation of "Centerness" maps \cite{wang2020learning}. After determining the positive candidate locations, other parameters in the labels also need to be assigned to them, see Section III-B for details. 

\begin{figure}[tbp]
	\vspace{-1.2em}
	\centering
	\epsfig{width=0.45\textwidth,file=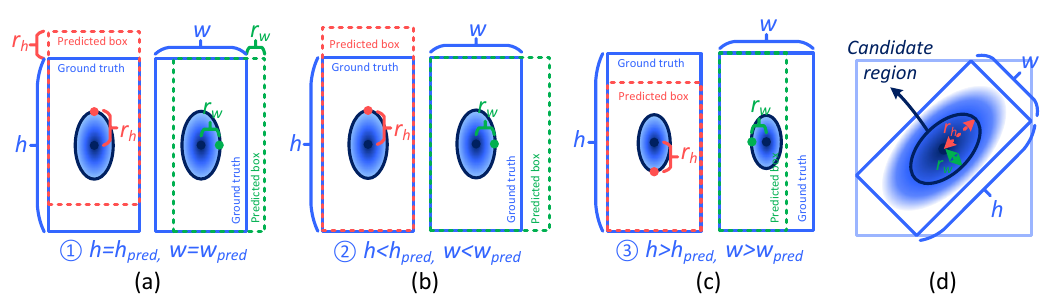}
	\vspace{-1em}
	\caption{The radiuses of the candidate region. (a) The case that the predicted box is as large as the ground truth. (b) The case that the predicted box is larger than the ground truth. (c) The case that the predicted box is smaller than the ground truth. (d) The radiuses of the candidate region.}
	\label{fig:5}
	\vspace{-1em}
\end{figure}

\textbf{3) Third, the spatial and scale extents of the candidate regions using the above strategy need to be carefully studied.} First, a bounding box centered at the Gaussian peak location (called C-BBox) is computed based on the assigned labels. Then, it is assumed that many bounding boxes of different sizes centered at the other Gaussian candidate locations are generated. At a location, if there exists a bounding box whose Intersection over Union (IoU) with the C-BBox is greater than the threshold ${T_{IoU}}$, this location is selected as a positive location. As shown in Fig.~\ref{fig:5}, these positive locations form a subset of the original Gaussian candidate locations (appearing as a smaller ellipse that is co-centered with the original Gaussian ellipse), and its semi-axis length is
\begin{equation}
	{r_i^c = \frac{{1 - {T_{IoU}}}}{2} \times {r_i},{\rm{  }}i = 1,2},
	\label{eq:7}
\end{equation}
where ${r_i},{\rm{ }}i = 1,2$, represents the semi-axis lengths of the original Gaussian ellipse. Then, the Gaussian boundary threshold in Algorithm~\ref{alg:1} is calculated from $r_i^c$. The purpose of the above reasoning is to explain the relationship between the Gaussian range and the IoU metric, and in practice it is not necessary to calculate the IoU during label assignment. Considering the versatility of multiple criteria, ${T_{IoU}}$ is set to 0.3, which is the same as many classic methods, like Faster R-CNN \cite{renFasterRCNNRealTime2017a} and YOLO \cite{redmonYOLOv3IncrementalImprovement2018}. 

In order to detect objects of different sizes on feature maps of different scales, objects’ OBBs with different sizes are assigned to feature maps with different down-sampling rates $strid{e_m} = {2^{m + 3}},{\rm{ }}m = 1,2,3$, as ground truth. The generated $\boldsymbol F_m$ from Algorithm~\ref{alg:1} of different scales are visualized in Fig.~\ref{fig:4} (d) (e) (f). To ensure that more than one positive candidate is generated on a certain scale after assignment, we set $\mathop {\max }\limits_i \left( {r_i^c} \right)/strid{e_m} \ge 1$, that is, $\mathop {\max }\limits_i \left( {2{r_i}} \right) \ge \frac{{2 \times strid{e_m}}}{{1 - {T_{IoU}}}}$ for each $m,{\rm{ }}m = 1,2,3$. Define the lengths of the four sides of an OBB as ${d_j},{\rm{ }}j = 1,2,3,4$, then $\mathop {\max }\limits_j \left( {{d_j}} \right) = \mathop {\max }\limits_i \left( {2{r_i}} \right) \ge \frac{{2 \times strid{e_m}}}{{1 - {T_{IoU}}}}$ because when calculating the diagonal matrix $\boldsymbol \Lambda$ in Eq.~\ref{eq:6} to generate the Gaussian ellipse, half the values of the length and width of the OBB are used as $s_1$ and $s_2$. Thus, introduce a hand-crafted hyperparameter $\tau  = 3$ to get two boundary values of the three assignment ranges, they are
\begin{equation}
	{rang{e_1} \! = \frac{{\tau  \times 2 \times strid{e_1}}}{{1 - {T_{IoU}}}},{\rm{  }}rang{e_2} \! = \frac{{\tau  \times 2 \times strid{e_3}}}{{1 - {T_{IoU}}}}}.
	\label{eq:8}
\end{equation}
When $\mathop {\max }\limits_j \left( {{d_j}} \right) \in \left( {1,rang{e_1}} \right]$, $\mathop {\max }\limits_j \left( {{d_j}} \right) \in \left( {rang{e_1},rang{e_2}} \right]$, and $\mathop {\max }\limits_j \left( {{d_j}} \right) \in \left( {rang{e_2},\sqrt 2 le{n^{img}}} \right]$, the object is assigned to feature maps with down sampling rates $stride{_1}$, $stride{_2}$, and $stride{_3}$, respectively. $le{n^{img}}$ represents the length or width of the image input to CNN. The hyperparameter $\tau$ is the only hand-crafted hyperparameter in the proposed GGHL. In Section IV, the setting of $\tau$ will be discussed later.

\vspace{-0.5em}
\subsection{ORC, OWAM, and CNN Model}
\begin{figure}[tp]
	\vspace{-1em}
	\centering
	\epsfig{width=0.35\textwidth,file=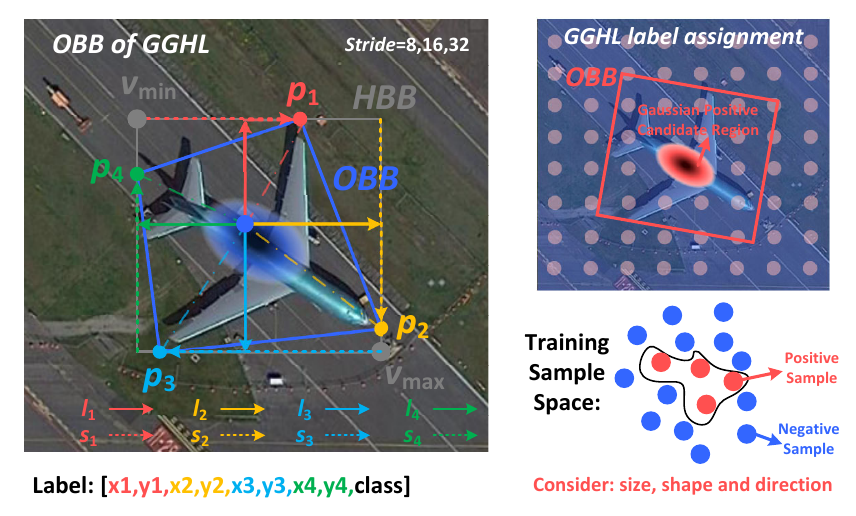}
	\vspace{-1em}
	\caption{The OBB representation of the proposed ORC.}
	\label{fig:6}
	\vspace{-1em}
\end{figure}

\textbf{1) Oriented-bounding-box representation component (ORC).} The proposed ORC is used to encode the ground truth labels and CNN’s predictions to represent objects in feature maps by their positive locations, OBBs, and categories. The existing OBB representation methods are divided into two main categories: angle-based and vertex-based. The angle-based methods, e.g., CenterMap \cite{wang2020learning}, only represent rotated rectangular bounding boxes, and the problem of periodicity and mutation in angle regression has been analyzed in GWD \cite{yang2021rethinking}. The vertex-based methods, such as Gliding Vertex \cite{xu2020gliding}, represent more other shapes of quadrilaterals, but do not account for the case where the vertices do not fall on the circumscribed HBB. Moreover, these OBB representations are based on anchor boxes, which are inflexible and depend on many anchor hyperparameters. The proposed ORC follows the simple principle of anchor-free 2-D Gaussian assignment, which directly represents the OBB using the horizontal and vertical components of the distances from each Gaussian candidate position to the four vertices of the OBB, as shown in Fig.~\ref{fig:6}. The proposed ORC is free from the dependence on the anchors in decoding the OBB and fits naturally with the proposed OLA. Moreover, the proposed ORC addresses the undiscussed case in Gliding Vertex \cite{xu2020gliding}, in which some vertices do not fall on the HBB.

The representation method of ORC is shown in Fig.~\ref{fig:6} and all the defined variables of ORC at the location ${\left( {x,y} \right)_m}$ are summarized in Table~\ref{table:1}. To represent an object in the feature map, first, the positive locations to detect the object are assigned. In the proposed OLA, the locations in the general Gaussian region are defined as the positive locations, while the other locations are defined as the negative locations. Thus, matrices $\boldsymbol {obj}{_m},{\rm{ }}m{\rm{ = 1,2,3}}$ are generated to represent the ground truth positive and negative locations, which are binary versions of the matrices $\boldsymbol{F}{_m}$. Let the component at location ${\left( {x,y} \right)_m}$ of $\boldsymbol {obj}{_m}$ be $ob{j_{x,y,m}}$. (For convenience, the subscripts $x,y,m$ are used in the following to indicate the variables at the location ${(x,y)_m}$.) If ${F_{x,y,m}} > 0$, $obj{_{x,y,m}} = 1$, and ${\left( {x,y} \right)_m}$ is a positive location. If ${F_{x,y,m}} = 0$, $obj{_{x,y,m}} = 0$, and ${\left( {x,y} \right)_m}$ is a negative location. In CNN, $\boldsymbol{\widehat {obj}}{_m},{\rm{ }}m = 1,2,3$, are generated to represent the estimation of $\boldsymbol{obj}{_m}$, whose component ${\widehat {obj}_{x,y,m}}$ at ${\left( {x,y} \right)_m}$ is in the range of $\left( {0,1} \right)$.
\begin{table}[tbp]
	\vspace{-1.5em}
	\centering
	%\small
	\renewcommand\arraystretch{1.5}
	\caption{{Summary of the definition of ORC variables at ${\left( {x,y} \right)_m}$}}
	\vspace{-0.5em}
	\label{table:1}
	\setlength{\tabcolsep}{1.5mm}{	
		\resizebox{0.48\textwidth}{!}{
		\begin{tabular}{c|m{4cm}<{\centering}|m{1.5cm}<{\centering}|m{1.5cm}<{\centering}}
				\hline\hline
				Variable &  Definition & Dimension & Value of Each Component \\ \hline
				$ob{j_{x,y,m}}$ & Ground truth representing that the location ${\left( {x,y} \right)_m}$ is positive or negative & \multirow{2}{1.5cm}{\centering Scalar} & 1 or 0 \\
				\cline{1,2,4}
				${\widehat {obj}_{x,y,m}}$ & Prediction score that 
				the location $ob{j_{x,y,m}}$ is positive
				 & ~ & $\left( {0,1} \right)$ \\
				\hline
				${\boldsymbol{l}_{x,y,m}}$ & \multirow{2}{3cm}{\centering Vector of the distances from $ob{j_{x,y,m}}$ to the HBB boundaries} & \multirow{2}{1.5cm}{\centering $1 \times 4$} & \multirow{2}{1.5cm}{\centering $\left[ {0,+\infty} \right)$} \\
				\cline{1}
				${\boldsymbol{\hat l}_{x,y,m}}$ & ~ & ~ & ~ \\
				\hline
				 ${\boldsymbol{s}_{x,y,m}}$ & \multirow{2}{4cm}{\centering Vector of the distances from the HBB vertices to the corresponding OBB vertices at ${\left( {x,y} \right)_m}$} & \multirow{2}{1.5cm}{\centering $1 \times 4$} & \multirow{2}{1.5cm}{\centering $\left[ {0,1} \right]$} \\
				\cline{1}
				${\boldsymbol{\hat s}_{x,y,m}}$ & ~ & ~ & ~ \\
				\hline
				${{ar}_{x,y,m}}$ & \multirow{2}{3cm}{\centering Area ratio of the HBB and OBB at ${\left( {x,y} \right)_m}$} & \multirow{2}{1.5cm}{\centering Scalar} & \multirow{2}{1.5cm}{\centering $\left[ {0,1} \right]$} \\
				\cline{1}
				${{\widehat {ar}}_{x,y,m}}$ & ~ & ~ & ~ \\
				\hline
				${\boldsymbol{obb}_{x,y,m}}$ & \multirow{2}{3cm}{\centering Vector representing the OBB at ${\left( {x,y} \right)_m}$} & \multirow{2}{1.5cm}{\centering $1 \times 9$} & \multirow{2}{1.5cm}{\centering $\left[ {0,1} \right]$} \\
				\cline{1}
				${\boldsymbol{\widehat {obj}}_{x,y,m}}$ & ~ & ~ & ~ \\
				\hline
				${\boldsymbol{cls}_{x,y,m}}$ & Ground truth vector of classification at ${\left( {x,y} \right)_m}$ & \multirow{2}{1.5cm}{\centering $1 \times nu{m_{cls}}$} & 1 or 0 \\
				\cline{1,2,4}
				${\boldsymbol{\widehat {cls}}_{x,y,m}}$ & Prediction vector 
				of classification at ${\left( {x,y} \right)_m}$ & ~ & $\left[ {0,1} \right)$ \\
				\hline\hline
		\end{tabular}}}	
	%\vspace{-2em}
\end{table}

\begin{figure}[tp]
	\vspace{-1.5em}
	\centering
	\epsfig{width=0.48\textwidth,file=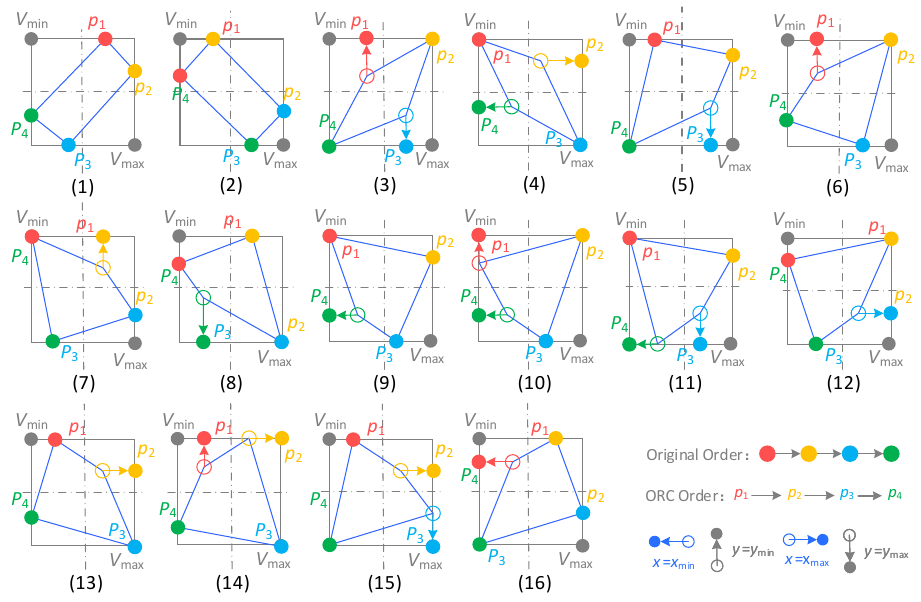}
	\vspace{-1em}
	\caption{Discussion of different cases of arbitrary convex quadrilateral represented by ORC. The colored letters $p_{1}$, $p_{2}$, $p_{3}$, and $p_{4}$ indicate the ordered convex quadrilateral vertices after ORC preprocessing.}
	\label{fig:6_A}
	\vspace{-1em}
\end{figure}

Second, when the positive locations are assigned, OBBs of different locations are represented for locating the objects more accurately. As shown in Fig.6, we use ${\boldsymbol{l}_{x,y,m}} = [{l_1},{l_2},{l_3},{l_4}]$, and ${\boldsymbol{s}_{x,y,m}} = [{s_1},{s_2},{s_3},{s_4}]$ to represent the OBB of an object at ${\left( {x,y} \right)_m}$. ${l_1},{l_2},{l_3},{l_4}$ are the distances from the location ${\left( {x,y} \right)_m}$ to the top, right, bottom, and left edges of the circumscribing horizontal bounding box (HBB) calculated from the ground truth coordinates. ${s_1},{s_2},{s_3},{s_4}$, are the distances from the vertices of the HBB to the corresponding vertices of the OBB. Note that $\boldsymbol{s}_{x,y,m}$ is normalized to the range of $\left[ {0,1} \right]$ by dividing by the corresponding side length of the HBB. Besides, as with Gliding Vertex \cite{xu2020gliding}, $a{r_{x,y,m}} \in \left[ {0,1} \right]$ are generated to represent the area ratio of the HBB and OBB. Thus, the OBB of the object at ${\left( {x,y} \right)_m}$ is represented by a $1 \times 9$-dimensional vector $\boldsymbol{obb}{_{x,y,m}} = \left[{{\boldsymbol{l}_{x,y,m}},{\boldsymbol{s}_{x,y,m}},ar{_{x,y,m}}} \right]$. Correspondingly, the CNN’s prediction of the OBB, $\boldsymbol{obb}{_{x,y,m}}$, at ${\left( {x,y} \right)_m}$ is represented as $\boldsymbol{\widehat {obb}}{_{x,y,m}} = \left[{{\boldsymbol{\hat l}_{x,y,m}},{\boldsymbol{\hat s}_{x,y,m}},\widehat {ar}{_{x,y,m}}} \right]$. 

Third, the object’s category is represented at each location. The ground truth classification at ${\left( {x,y} \right)_m}$ is represented as a $1 \times nu{m_{cls}}$-dimensional one-hot vector $\boldsymbol{cls}{_{x,y,m}} = \left[ {cls_{x,y,m}^{\left( 1 \right)}, \cdots ,cls_{x,y,m}^{\left( {num{_{cls}}} \right)}} \right]$, where $nu{m_{cls}}$ denotes the number of categories. Let the $c$th component of $\boldsymbol{cls}{_{x,y,m}}$ be $cls_{x,y,m}^{\left( c \right)} \in \left\{ {0,1} \right\}$, $c \in A = \left\{ {1,2, \cdots ,nu{m_{cls}}} \right\}$. If the object at location ${\left( {x,y} \right)_m}$ belongs to the cth category, $cls_{x,y,m}^{\left( c \right)} = 1$; otherwise, $cls_{x,y,m}^{\left( c \right)} = 0$. Correspondingly, the CNN’s prediction of $\boldsymbol{cls}{_{x,y,m}}$ is represented as $\boldsymbol{\widehat {cls}}{_{x,y,m}}=\left[ {\widehat {cls}_{x,y,m}^{\left( 1 \right)}, \cdots ,\widehat {cls}_{x,y,m}^{\left( {nu{m_{cls}}} \right)}} \right]$, the component $\widehat {cls}_{x,y,m}^{\left( c \right)} \in \left( {0,1} \right)$ of which represents the probability that the object belongs to the $c$th category. 

\textbf{2) Refined approximation of OBBs.} Furthermore, not all convex quadrilaterals are directly represented by the ideal ORC drawn as shown in Fig.~\ref{fig:6_A}. The cases that the vertices do not fall on the HBB and the implicit ordering of vertices in the ORC need to be discussed. These are not fully considered in vertex-based methods, such as Gliding Vertex \cite{xu2020gliding}. They cope with these cases by converting the quadrilateral to its minimal outer rectangle. But such large-scale one-size-fits-all approximate conversions introduce large errors. And vertex-based methods only represent rotated rectangles and not arbitrary convex quadrilaterals. In response, we generalize and discuss the problem more comprehensively by generalizing the ORC representation and vertex ordering of arbitrary convex quadrilaterals to 16 cases. For their interpretations, schematic diagrams are more intuitive and easier to understand than words, so a summary of these cases is illustrated in Fig.~\ref{fig:6_A}. According to this refined approximation (RA), it is only necessary to make different types of approximations with as small error as possible for few convex quadrilaterals to represent arbitrary convex quadrilaterals and to obtain implicitly ordered vertices. Based on the statistics of more than two million OBBs in the DOTA dataset \cite{xiaDOTALargeScaleDataset2018}, only 4.79\% of the OBBs need to be approximated. The error introduced by using the minimum outer rectangle approximation for all the “difficult” convex quadrilaterals (counted by pixel areas) is more than twice as large as the one in the proposed method. Due to the space limitation, a more detailed algorithm is available in our open-source codes (https://github.com/Shank2358/GGHL). 

After the above variables are obtained and represented, the CNN training process is to make the CNN’s predictions approach the ground truth values, i.e., minimizing the loss in Eq.~\ref{eq:19}, which will be described later.

\begin{figure}[bp]
	\vspace{-2em}
	\centering
	\epsfig{width=0.42\textwidth,file=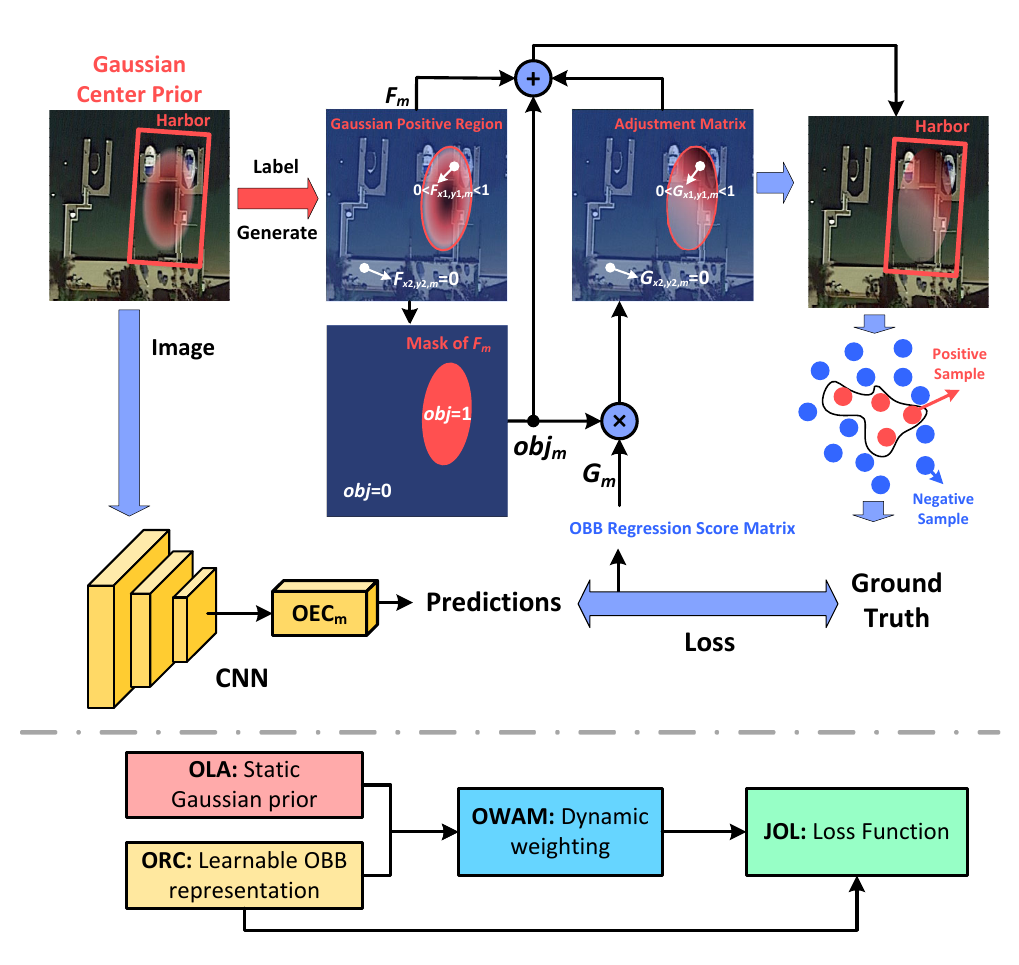}
	\vspace{-1em}
	\caption{The object-adaptive weight adjustment mechanism (OWAM) based on Gaussian center prior (GCP) weight and OBB shape regression score.}
	\label{fig:7}
	\vspace{-0.5em}
\end{figure}

\textbf{3) Object-adaptive weight adjustment mechanism (OWAM).} Generally, after generating an elliptical Gaussian candidate region and assigning labels to all the locations in this region, as shown in Fig.~\ref{fig:6}, the value of $f\left( {x,y} \right)$ is used to weight a location of the candidate region when calculating the location loss. However, some objects like harbors in the remote sensing datasets as shown in Fig.~\ref{fig:7} do not conform to the Gaussian center prior. Therefore, it is not appropriate to use Gaussian weight directly. This has not been considered by the existing Gaussian-center-prior methods, such as CenterNet \cite{zhou2019objects}, BBAVectors \cite{yi2020oriented}, DRN \cite{pan2020dynamic}, O$^2$-DNet \cite{wei2020oriented}, and loss functions like GWD \cite{yang2021rethinking}. In the field of horizontal object detection, AutoAssign \cite{zhu2020autoassign} and IQDet \cite{ma2021iqdet} employed the adaptive weight adjustment with success. OWAM borrows this idea and extends it to oriented object detection. Compared to the existing methods, benefitting from using general Gaussian PDFs defined in OLA as prior weights, the proposed OWAM represents translation, rotation, and scaling for the arbitrary-oriented object. This Gaussian prior is for each individual while not for each category designed in AutoAssign \cite{zhu2020autoassign}. Besides, as mentioned before, using general Gaussian instead of “Rotated-Centerness-like” mechanisms to learn "Objectness" avoids the possible interpolation approximation problem. Based on this prior, the weights are dynamically adjusted by the orientation correlation variables $\boldsymbol{s}_{x,y,m}$ and ${ar}_{x,y,m}$ designed in ORC during the CNN training. That is, OWAM combines the static prior in OLA with the dynamically learnable OBB representation in ORC to re-weight the assigned candidates.

In OWAM, the higher weights are assigned to the key positive locations of an object learned by CNN, while not always the center point of an object that is used by ${\boldsymbol{F}_m},{\rm{ }}m = 1,2,3$. As shown in Fig.~\ref{fig:7}, weight adjustment matrices ${\boldsymbol{G}_m},{\rm{ }}m = 1,2,3$ are introduced to adaptively adjust the weight of each Gaussian region of ${\boldsymbol{F}_m}$, according to object’s shape. Note that matrices ${\boldsymbol{G}_m}$ are calculated based on the proposed ORCs of CNN using the OBB regression loss described below, which reflects the OBB shape prediction scores at positive locations. The value of each component in ${\boldsymbol{G}_m}$ is in the range of $\left( {0,1} \right)$. More specifically, based on the above OBB encoding process, the OBB regression loss at each positive location ${\left( {x,y} \right)_m},{\rm{ }}m = 1,2,3$, is
\begin{equation}
	\begin{array}{l}
	\!	Loss \! \left( {\! \boldsymbol{obb}{_{x,y,m}},\!{{\boldsymbol{\widehat {obb}}}_{x,y,m}}} \! \right) \! = \! 1 \! - \! {GIoU} \! \left( {\!{\boldsymbol{l}\!_{x,y,m}},\!{{\boldsymbol{\hat l}}\!_{x,y,m}}} \! \right)\\
		+ \sum\limits_{k = 1}^4 {{{\left( {s_{x,y,m}^{\left( k \right)} - \hat s_{x,y,m}^{\left( k \right)}} \right)}^2}}  + {\left( {a{r_{x,y,m}} - {{\widehat {ar}}_{x,y,m}}} \right)^2}.
	\end{array}
	\label{eq:9}
\end{equation}
The loss function in Eq.~\ref{eq:9} is obtained from maximizing the likelihood function of the parameters to estimate. It is explained in Appendix A-1). $GIoU\left(  \cdot  \right)$ function \cite{rezatofighiGeneralizedIntersectionUnion2019} is an improved IoU for training, the calculation of which is given in Appendix B. $\hat s_{x,y,m}^{\left( k \right)}$ is the $k$th component of $1 \times 4$-dimensional vector ${\boldsymbol{\hat s}_{x,y,m}}$, and $s_{x,y,m}^{\left( k \right)}$ is the $k$th component of $1 \times 4$-dimensional vector ${\boldsymbol{s}_{x,y,m}}$. Therefore, the output of $Loss \left( {\boldsymbol{\hat obb}{_{x,y,m}},{{\boldsymbol{\widehat {obb}}}_{x,y,m}}} \right)$ is a scalar greater than or equal to 0 at location ${\left( {x,y} \right)_m}$. The smaller its value is, the more accurate the prediction of OBB is.

Thus, ${e^{ - Loss \left( {\boldsymbol{obb}{_{x,y,m}},{{\boldsymbol{\widehat {obb}}}_{x,y,m}}} \right)}}$ is in the range of $\left( {0,1} \right]$. Let the value of ${\boldsymbol{G}_m}$ at location ${\left( {x,y} \right)_m},{\rm{ }}m = 1,2,3,$ be
\begin{equation}
	\begin{array}{l}
	{G_{x,y,m}} = {e^{ - Loss \left( {\boldsymbol{obb}{_{x,y,m}},{{\boldsymbol{\widehat {obb}}}_{x,y,m}}} \right)}}.
	\end{array}
	\label{eq:10}
\end{equation}
The larger its value is, the more accurate the prediction of OBB is. Then, ${\boldsymbol{G}_m}$ is adaptively adjusted according to the shape of the objects to be predicted by the CNN training. So, the weight of ${\left( {x,y} \right)_m}$ is changed from ${f_m}\left( {x,y} \right)$ to 
\begin{equation}
	\begin{array}{l}
		weight_{x,y,m}^{obb} = 1 - obj{_{x,y,m}} + \vartheta {f_m}\left( {x,y} \right) \\+ \left(1-\vartheta\right) {G_{x,y,m}} \times ob{j_{x,y,m}},
	\end{array}
	\label{eq:11}
\end{equation}
where scalar $weight_{x,y,m}^{obb} \in \left( {0,1} \right]$ represents the object-adaptive weight at ${\left( {x,y} \right)_m}$. $\vartheta  \in \left( {0,1} \right)$ denotes the weighting factor. $\vartheta=1$ means that the weights are completely dependent on the Gaussian prior, and $\vartheta \ne 0$. If ${\left( {x,y} \right)_m}$ is a negative location, then $weight_{x,y,m}^{obb} = 1$; otherwise, $weight_{x,y,m}^{obb} \in \left( {0,1} \right)$. Finally, ${W_m} \times {H_m}$-dimensional CNN-learnable weight matrices $\boldsymbol{weight}_m^{obb}$ composed of $weight_{x,y,m}^{obb}$ are generated, where ${W_m} = {W \mathord{\left/	{\vphantom {W {strid{e_m}}}} \right.	\kern-\nulldelimiterspace} {stride{_m}}}$ and ${H_m} = {H \mathord{\left/	{\vphantom {H {stride{_m}}}} \right.	\kern-\nulldelimiterspace} {stride{_m}}}$ represent the width and height of the feature map of scale   respectively. $\boldsymbol{weight}_m^{obb},{\rm{ }}m = 1,2,3,$ weight different locations dynamically to fit different object shapes in different scales. Besides, in the above process, there is no need to change the label assignments designed in Section III-A. Compared with the scheme of using CNN to predict the weights directly, this scheme makes the adjustment converge faster and more stable based on ${\boldsymbol{F}_m}$, during the CNN training.  

\textbf{4) CNN model.} Many methods use large and complicated CNN models to pursue high accuracy, which may be too complex in practical applications. For computational simplicity, the CNN model of the proposed GGHL chooses a straightforward and practical structure for its versatility and ease of use, which can more precisely reflect the effectiveness of the proposed GGHL. The designed CNN model is mainly composed of three parts: backbone network, feature pyramid network (FPN) \cite{linFeaturePyramidNetworks2017a}, and detection head composed of ORCs, which are represented by red, blue and yellow in Fig. 4 (b), respectively. Considering that the object scale varies greatly in remote sensing scenes, spatial pyramid pooling (SPP) \cite{huang2020dc} is introduced in FPN to fuse multiscale features and expand the receptive field. In addition, the DropBlock \cite{ghiasi2018dropblock} is used to improve the generalization ability of CNN, which does not bring additional computational complexity. The detection head uses a very light two-layer convolution structure, unlike the heavy convolution layers of RetinaNet \cite{linFocalLossDense2017}.

\vspace{-0.5em}
\subsection{Joint-optimization Loss (JOL)}
First, the joint PDF of the P\&N location detection, OBB regression, and object classification at each location of feature maps is provided. Second, an area normalization and loss re-weighting mechanism is designed for adaptively adjusting the weight of loss at different locations. Finally, the maximum likelihood estimation (MLE) is used to obtain the total joint-optimization function. Besides, the CNN predictions in the inference stage are explained.

\textbf{1) The joint PDF of the positive or negative location detection, OBB regression, and object classification.} Use $\boldsymbol{loc}{_{x,y,m}} = \left[ {obj{_{x,y,m}},\boldsymbol{obb}{_{x,y,m}},\boldsymbol{cls}{_{x,y,m}}} \right]$ to represent the ground truth of object detection at location ${\left( {x,y} \right)_m}$. For the CNN model, let $\boldsymbol{\theta}{^{loc}_{x,y,m}} = \left[ {\theta{^{obj}_{x,y,m}},\boldsymbol{\theta}{^{obb}_{x,y,m}},\boldsymbol{\theta}{^{cls}_{x,y,m}}} \right]$ be the CNN parameters used for object detection at the location ${\left( {x,y} \right)_m}$. $\theta{^{obj}_{x,y,m}}$, $\boldsymbol{\theta}{^{obb}_{x,y,m}}$, and $\boldsymbol{\theta}{^{cls}_{x,y,m}}$ are the parameters used for the positive or negative location detection, OBB regression, and object classification, respectively. Similarly, let $\boldsymbol{x}{^{loc}_{x,y,m}} = \left[ {x{^{obj}_{x,y,m}},\boldsymbol{x}{^{obb}_{x,y,m}},\boldsymbol{x}{^{cls}_{x,y,m}}} \right]$ be the input features of the prediction layers of CNN at ${\left( {x,y} \right)_m}$, which are extracted by the hidden layers of CNN. 

Then, define the predictions of CNN as $\boldsymbol{\widehat{loc}}{_{x,y,m}} = \left[ {\widehat{obj}{_{x,y,m}},\boldsymbol{\widehat{obb}}{_{x,y,m}},\boldsymbol{\widehat{cls}}{_{x,y,m}}} \right]$, which is generated by $\boldsymbol{\theta}{^{loc}_{x,y,m}}$ and $\boldsymbol{x}{^{loc}_{x,y,m}}$. Specifically, for the positive or negative location detection, define $n{n^{obj}}\left(  \cdot  \right)$ as a deterministic function with Sigmoid activation related to CNN, then,
\begin{equation}
	\begin{array}{l}
{\widehat {obj}_{x,y,m}} = n{n^{obj}}\left( {x_{x,y,m}^{obj},\theta _{x,y,m}^{obj}} \right).
	\end{array}
	\label{eq:12}
\end{equation}
The estimation ${\widehat {obj}_{x,y,m}}$ is in the range of $\left( {0,1} \right)$, which represents the classification score that the location ${\left( {x,y} \right)_m}$ is positive. The larger the ${\widehat {obj}_{x,y,m}}$ is, the more likely ${\left( {x,y} \right)_m}$ is to be a positive location.

For the OBB regression, define $n{n^{obb}}\left(  \cdot  \right)$ as a deterministic regression function related to the CNN, which uses the linear activation function [1]. Note that in the CNN training stage, the ground truth of positive and negative location detection, i.e., $ob{j_{x,y,m}} \in \left\{ {0,1} \right\}$, is given. The estimation of OBB is only carried out at the positive locations \cite{renFasterRCNNRealTime2017a}:
\begin{equation}
	\begin{array}{l}
	{\boldsymbol{\widehat {obb}}_{x,y,m}} = ob{j_{x,y,m}} \times nn{^{obb}}\left( {\boldsymbol{x}_{x,y,m}^{obb},\boldsymbol{\theta}_{x,y,m}^{obb}} \right),
	\end{array}
	\label{eq:13}
\end{equation}
which is used in the joint PDF and loss function. While in the CNN inference stage, $obj{_{x,y,m}}$ is unknown, but ${\widehat {obj}_{x,y,m}}$ has been obtained after the training, which will be explained in detail in Sub-section 4).

For the object classification, the parameters $\boldsymbol{\theta}_{x,y,m}^{cls} = \left[ {\theta {{_{x,y,m}^{cls}}^{\left( 1 \right)}}, \cdots ,\theta {{_{x,y,m}^{cls}}^{\left( {nu{m_{cls}}} \right)}}} \right]$, and the input features $\boldsymbol{x}_{x,y,m}^{cls} = \left[ {x{{_{x,y,m}^{cls}}^{\left( 1 \right)}}, \cdots ,x{{_{x,y,m}^{cls}}^{\left( {nu{m_{cls}}} \right)}}} \right]$. Define $n{n^{cls}}\left(  \cdot  \right)$ as a deterministic function with Sigmoid activation related to the CNN. Note that in the CNN training stage, $obj{_{x,y,m}}$ and $\boldsymbol{obb}{_{x,y,m}}$ are given, and ${G_{x,y,m}}$ is calculated after $\boldsymbol{\widehat {obb}}{_{x,y,m}}$ is predicted by the CNN. In this stage, the estimation of classification is only carried out at the positive locations, i.e. $obj{_{x,y,m}} = 1$ \cite{redmonYouOnlyLook2016a}. In the existing methods like \cite{redmonYOLOv3IncrementalImprovement2018,linFocalLossDense2017,tian2019fcos,yangSCRDetMoreRobust2019,xu2020gliding}, the classification score is usually learned independently by CNN. While in the proposed GGHL, ${G_{x,y,m}} \in \left( {0,1} \right]$ is multiplied to the classification score, which makes the classification score also affected by the OBB regression score. Thus, the estimation that object belongs to the $c$th category is
\begin{equation}
	\begin{array}{l}
		\widehat {cls}_{x,y,m}^{\left( c \right)} = ob{j_{x,y,m}} \times {G_{x,y,m}} \\ 
		\times nn{^{cls}}\left( {x{{_{x,y,m}^{cls}}^{\left( c \right)}},\theta {{_{x,y,m}^{cls}}^{\left( c \right)}}} \right),
	\end{array}
	\label{eq:14}
\end{equation}
where ${G_{x,y,m}}$ is given in Eq.~\ref{eq:10}. The estimation $\widehat {cls}_{x,y,m}^{\left( c \right)}$ activated by Sigmoid function is in the range of $\left( {0,1} \right)$. The larger the $\widehat {cls}_{x,y,m}^{\left( c \right)}$ is, the more likely the object at ${\left( {x,y} \right)_m}$ is to belong to the $c$th category. Therefore, the classification sub-task is affected by the OBB regression error. In the training process, in order to obtain a higher classification accuracy, the model parameters will be jointly adjusted to approach the optimal results of not only the classification sub-task but also the OBB regression task. Thus, when $\boldsymbol{\theta}_{x,y,m}^{loc}$ and $\boldsymbol{x}_{x,y,m}^{loc}$ are given, the joint PDF of the positive or negative location detection, OBB regression, and object classification is
\begin{equation}
	\begin{array}{l}
		p\left( {\boldsymbol{loc}{_{x,y,m}}\left| {\boldsymbol{x}_{x,y,m}^{loc},\boldsymbol{\theta} _{x,y,m}^{loc}} \right.} \right) \\
		= p\left( {obj{_{x,y,m}}\left| {x_{x,y,m}^{obj}} \right.;\theta _{x,y,m}^{obj}} \right)\\
		\times p\left( {\boldsymbol{obb}{_{x,y,m}}\left| {obj{_{x,y,m}};\boldsymbol{x}_{x,y,m}^{obb}} \right.;\boldsymbol{\theta}_{x,y,m}^{obb}} \right)\\
		\times p(cls_{x,y,m}^{\left( 1 \right)} \cdots cls_{x,y,m}^{\left( {nu{m_{cls}}} \right)}\left| {\boldsymbol{obb}{_{x,y,m}};ob{j_{x,y,m}};} \right.\\
		x_{x,y,m}^{\left( 1 \right)} \cdots x_{x,y,m}^{\left( {nu{m_{cls}}} \right)};{\theta {_{x,y,m}^{cls}}^{\left( 1 \right)}}, \cdots ,{\theta {_{x,y,m}^{cls}}^{\left( {nu{m_{cls}}} \right)}}),
	\end{array}
	\label{eq:15}
\end{equation}
which is derived in Appendix A. The PDF of the error of the OBB regression, which is assumed to obey an i.i.d. Gaussian distribution with a mean of 0 and variance ${\sigma ^2}$. 

\textbf{2) Area normalization and loss adaptive re-weighting.} Because CNN prefers to learn the object with a larger Gaussian region generated by the proposed OLA, i.e., with more positive locations, an area normalization factor ${\xi _{x,y,m}}$ at ${\left( {x,y} \right)_m}$ that decreases with increasing Gaussian candidate areas is considered. The statistics of OLA assignment for many AOOD datasets, such as DOTA \cite{xiaDOTALargeScaleDataset2018}, show that the number of objects and the area of the assigned candidate region (number of pixels) exhibit a decreasing trend from fast to slow. Therefore, the reciprocal form of the logarithm is chosen to design variables ${\xi _{x,y,m}}$ so that the variation trend approximates the distribution described above with a lower bound. According to Eq.~\ref{eq:8}, the theoretical maximum value of candidate area is ${\left( {{{le{n^{img}} \times \left( {1 - {T_{IoU}}} \right)} \mathord{\left/ {\vphantom {{le{n^{img}} \times \left( {1 - {T_{IoU}}} \right)} {32}}} \right. \kern-\nulldelimiterspace} {32}}} \right)^2}$, and the variation of this value is still large, so its square root is taken. To ensure that the denominator is not 0, 1 is added to the log of the denominator. To make the maximum value be 1, the numerator is log2. The designed area normalization variable is
\begin{equation}
	\begin{array}{l}
        \xi_{x,y,m} = \displaystyle{\frac{{\log 2}}{{\log \left( {1 + \sqrt {are{a_{x,y,m}}} } \right)}}},
	\end{array}
	\label{eq:16}
\end{equation}
where $area{_{x,y,m}}$ denotes the area of positive region and is always no less than 1. The normalization weight is in the range of $\left( {0,1} \right]$.

In JOL, to make the detection of positive and negative affected by the object’s shape, $weight_{x,y,m}^{obb}$ designed in Eq.~\ref{eq:11} is used to adaptively weight the location loss according to the error of OBB regression, i.e., the error of object’s shape prediction. Besides, to impose classification effects on regression, the weight $weight_{x,y,m}^{cls}$ is designed to weight the OBB regression loss after ${G_{x,y,m}}$ is obtained in the total loss,
\begin{equation}
\begin{array}{l}
	weight_{x,y,m}^{cls} = 1 - obj{_{x,y,m}} + \vartheta {f_m}\left( {x,y} \right) + \\ \left(1-\vartheta\right) nn{^{cls}}\left( {x{{_{x,y,m}^{cls}}^{\left( c \right)}},\theta {{_{x,y,m}^{cls}}^{\left( c \right)}}} \right) ,
\end{array}
	\label{eq:17}
\end{equation}
where the ground truth category is the $c$th category. Similar to Eq.~\ref{eq:11}, $weight_{x,y,m}^{cls}$ is also in the range $\left( {0,1} \right]$. $1 - obj{_{x,y,m}}$ is used to make the non-object part of the weights equal to 1. Here, $weight_{x,y,m}^{obb}$ and $weight_{x,y,m}^{cls}$ do not perform the gradient backpropagation during training. In GGHL, taking $\vartheta=0.5$ obtains equal contributions from the prior weights and adjusted values, which may not be optimal but is simplest.

\textbf{3) Total joint-optimization function.} After caonsidering ${\xi _{x,y,m}}$ and $weight_{x,y,m}^{obb}$, and introducing the Focal Loss \cite{linFocalLossDense2017}, from the LF of Eq.~\ref{eq:15} and using the MLE, the total loss of all the locations in feature maps is obatined, which is 
\begin{equation}
\begin{array}{l}
	\! Loss{_{total}} \\ = Loss \left( \boldsymbol{obj}, \boldsymbol{\widehat {obj}} \right)
	\times weight_{x,y,m}^{obb} \times {\xi _{x,y,m}} \\
	+ Loss \left( {\! \boldsymbol{obb},{{\boldsymbol{\widehat {obb}}}}} \right) \times weight_{x,y,m}^{cls} \times {\xi _{x,y,m}}\\
	+ Loss \left( {\! \boldsymbol{cls},\!{{\boldsymbol{\widehat {cls}}}}} \right) \times {\xi _{x,y,m}}.
\end{array}
	\label{eq:18}
\end{equation}
In the total loss, the loss of P\&N location detection is
\begin{equation}
	\begin{array}{l}
		Loss \left( \boldsymbol{obj}, \boldsymbol{\widehat {obj}} \right)= \\
		\!- \!\sum\limits_{\scriptstyle x,y \in F{M_m}\atop
			\scriptstyle m = 1,2,3} \!{{(1 \!- {{\widehat {obj}}_{x,y,m}})}^\gamma } \! \log \left( {obj{_{x,y,m}}} \right) \\ 
		\!- \!\sum\limits_{\scriptstyle x,y \in F{M_m}\atop
			\scriptstyle m = 1,2,3} {{{({{\widehat {obj}}_{x,y,m}})}^\gamma }\log \left( {1 - ob{j_{x,y,m}}} \right)},
	\end{array}
	\label{eq:18_A}
\end{equation}
where $F{M_m}$ represents the feature maps in scales $m,{\rm{ }}m = 1,2,3$. $\gamma$ is the hyperparameter of Focal Loss \cite{linFocalLossDense2017}, which is set to 2 as \cite{linFocalLossDense2017}. In JOL, the loss of P\&N location detection is separated from the classification loss so that the imbalance of P\&N samples will not affect the classification task. The OBB regression loss is
\begin{equation}
	\begin{array}{l}
		Loss \left( {\! \boldsymbol{obb},{{\boldsymbol{\widehat {obb}}}}} \right) = \\ \sum\limits_{\tiny{\! \scriptstyle \! x,y \in F{M_m}\atop
			\scriptstyle m \! = 1,2,3}} \! Loss \left( {\! \boldsymbol{obb}{_{x,y,m}},\!{{\boldsymbol{\widehat {obb}}}_{x,y,m}}} \! \right),
	\end{array}
	\label{eq:18_B}
\end{equation}
and the classification loss is
\begin{equation}
	\begin{array}{l}
	Loss \left( {\! \boldsymbol{cls},\!{{\boldsymbol{\widehat {cls}}}}} \right) = \\
	- \sum\limits_{\scriptstyle x,y \in F{M_m}\atop
	\scriptstyle m = 1,2,3} {\sum\limits_{c = 1}^{nu{m_{cls}}} {(cls_{x,y,m}^{\left( c \right)}\log \left( {\widehat {cls}_{x,y,m}^{\left( c \right)}} \right)} } \\
+ \left( {1 - cls_{x,y,m}^{\left( c \right)}} \right)\log \left( {1 - \widehat {cls}_{x,y,m}^{\left( c \right)}} \right)),
\end{array}
\label{eq:18_C}
\end{equation}
where the classification estimation $\widehat {cls}_{x,y,m}^{\left( c \right)}$ defined by Eq.~\ref{eq:14} is associated with the OBB regression result $G_{x,y,m}$. This is different from the ordinary loss scheme in which independent regression loss and classification loss are added together.

\textbf{4) The CNN predictions in the inference stage.} Note that ${\boldsymbol{\widehat {obb}}_{x,y,m}}$ and ${\boldsymbol{\widehat {cls}}_{x,y,m}}$ in the CNN training stage and inference stage are different. In the CNN inference stage, $obj{_{x,y,m}}$ is unknown, after ${\widehat {obj}_{x,y,m}}$ is obtained. If ${\widehat {obj}_{x,y,m}}$ is larger than the threshold, which is given by the benchmarks of different datasets, the location is predicted as a positive location,
\begin{equation}
{\boldsymbol{\widehat {obb}}_{x,y,m}} = n{n^{obb}}\left( {\boldsymbol{x}_{x,y,m}^{obb},\boldsymbol{\theta} {{_{x,y,m}^{obb}}^ * }} \right),
	\label{eq:19}
\end{equation}
\begin{equation}
	{\boldsymbol{\widehat {cls}}_{x,y,m}} = n{n^{cls}}\left( {\boldsymbol{x}_{x,y,m}^{cls},\boldsymbol{\theta} {{_{x,y,m}^{cls}}^ * }} \right),
	\label{eq:20}
\end{equation}
where $\boldsymbol{\theta} {{_{x,y,m}^{obb}}^ * }$  and $\boldsymbol{\theta} {{_{x,y,m}^{cls}}^ * }$ represent the optimal parameters obtained from the CNN training for OBB regression and object classification, respectively. If ${\widehat {obj}_{x,y,m}}$ is less than the threshold, the location is predicted as a negative location, and the OBB regression and object classification are not performed.

\section{Experiments and Discussions}
In this section, experiments on public AOOD datasets are conducted to verify the effectiveness of the proposed GGHL. First, the experimental conditions are explained. Secondly, the ablation experiments are conducted, the effectiveness of each components is analyzed, and the results are discussed. Furthermore, the proposed GGHL is used to replace the label assignment strategy of other mainstream AOOD methods to evaluate its versatility. Besides, the lightweight AOOD model LO-Det \cite{huang2021lo} is improved by the proposed GGHL, and its performance is evaluated on embedded platforms to verify the application friendliness. Third, comparative experiments on several public datasets of different scenes are evaluated to compare the performance of the proposed GGHL with the state-of-the-art methods.

\vspace{-0.5em}
\subsection{Experimental Conditions}
\textbf{1) Experimental platforms.} All the experiments were implemented on a computer with an AMD 5950X CPU, 128 GB of memory, and two NVIDIA GeForce RTX 3090 GPU (2$\times$24GB). Besides, in order to evaluate the application friendliness of the proposed GGHL, the embedded devices NVIDIA Jetson AGX Xavier and NVIDIA Jetson TX2 were also used for application experiments.

\textbf{2) Datasets.} In order to evaluate the performance of the proposed GGHL fully, multiple public datasets of different scenes and different image types are employed.

a) DOTA \cite{xiaDOTALargeScaleDataset2018} is currently the largest AOOD dataset containing 2806 aerial images from 800 × 800 pixels to 4000 × 4000 pixels, in which more than 188,000 objects falling into 15 categories are annotated. Due to the huge size, these images are usually \cite{huang2021lo} cropped into sub-images of 800×800 pixels with an overlap of 200 pixels. In addition, the multi-scale cropping (MSC) strategy is used like many recently proposed AOOD methods \cite{ming2021optimization,yang2021rethinking,han2021redet}. For MSC, the original images are scaled to [0.5; 1.0; 1.5] and then cropped into patches of size 800 $\times$ 800. The categories of the objects in DOTA are: Plane (PL), Baseball diamond (BD), Bridge (BR), Ground field track (GFT), Small vehicle (SV), Large vehicle (LV), Ship (SH), Tennis court (TC), Basketball court (BC), Storage tank (ST), Soccer-ball field (SBF), Roundabout (RA), Harbor (HA), Swimming pool (SP), and Helicopter (HC). DOTAv2.0 \cite{ding2021object} further expands three categories of objects, i.e., container-crane, airport, and helipad, based on DOTAv1.0 dataset. DOTAv2.0 contains 11,268 images and 1,793,658 instances, which is currently the largest AOOD dataset.
 
b) SKU110-R \cite{pan2020dynamic} is a dense oriented commodity detection dataset. The images are collected from thousands of supermarket stores. It is an extension of the original SKU110K dataset containing 1,733,678 instances. The number of images in training set, validation set and test set are 57533, 4116, and 20552, respectively.

c) SSDD+ \cite{li2017ship} is a polarized synthetic aperture radar (SAR) image dataset. It has 1,160 ship images including 2456 instances collected by RadarSat-2, TerraSARX, and Sentinel-1 sensors under different sea conditions. The polarization modes contain HH, HV, VV, and VH. The ratio of training set , validation set and testing set is 7:1:2.

\textbf{3) Evaluation metrics.} The mean Average Precision (mAP) with IoU threshold = 0.5, the widely used metric in OD tasks is adopted for evaluating the detection accuracy. The average precision of each category is AP. AP with an IoU threshold of 0.3 is represented as AP@0.3. The inference frames per second (fps) are used to evaluate the detection speed. The floating point of operations (FLOPs) is used to evaluate the computational complexity of the model. The memory occupied by parameters is used to evaluate the model size.

\textbf{4) Implementation details.} To compare the proposed GGHL with state-of-the-art methods fairly, training hyperparameters are set to be the same as the methods compared. The initial learning rate is set as 2×10$^{-4}$. The final learning rate is 1×10$^{-6}$, and the SGD strategy is adopted. Weight decay is 5×10$^{-4}$, and momentum is 0.9. The maximum training epoch is 36. The confidence threshold is 0.2, and the non-maximum suppression (NMS) threshold is 0.45. Data augmentation strategies including mixup, random cropping, and random flipping are used.

\textbf{5) Baseline \& Comparative methods.} An OD model usually consists of CNN parts and non-CNN parts. In order to evaluate the performance of each component proposed in GGHL, two models with the same CNN structure are constructed as baselines. Among these two Vanilla models, the one adopting the anchor-based label assignment strategy is called Vanilla-AB, while the other one using the anchor-free standard-Gaussian-based label assignment strategy is called Vanilla-AF. The two CNNs are only slightly different in the number of feature maps in the output layer. They both employ the OBB representation method of Gliding Vertex \cite{xu2020gliding} (called Vanilla-Head in the experiments) with static candidate region, and the loss function with additive paradigm.

In order to compare and analyze the performance of the proposed GGHL more comprehensively, many state-of-the-art AOOD methods are selected for comparison, such as SCRDet \cite{yangSCRDetMoreRobust2019}, Gliding Vertex \cite{xu2020gliding}, RIL \cite{ming2021optimization}, etc. Moreover, some popular anchor-free models like CenterNet \cite{zhou2019objects} and FCOS \cite{tian2019fcos} and the latest AOOD models like NPMMR-Det \cite{9364888} and LO-Det \cite{huang2021lo} are also adopted as the baseline to evaluate the versatility of the proposed GGHL.

\vspace{-0.5em}
\subsection{Ablation Experiments and Discussions}
\textbf{1) Ablation experiments of each component.} Ablation results of each component on the DOTA dataset are listed in Table~\ref{table:2}. The more detailed experimental results for each category are listed in Table~\ref{table:3}. First, an anchor-based detector Vanilla-AB is constructed, and the effect of the widely-used MSC data augmentation strategy \cite{han2021redet} is evaluated. From the experimental results, it can be seen that the MSC strategy increases mAP by 2.09 on the DOTA dataset. It can be observed from Table~\ref{table:2} that the average precision (AP) improvement of using MSC is more obvious for extreme scale objects, including large-scale objects, such as GFT and SF, and small-scale objects like SV.
\begin{table*}[htbp]
	\centering
	\vspace{-1.5em}
	\renewcommand\arraystretch{1.3}
	\setlength{\tabcolsep}{1.5mm}{
		\caption{\label{table:2}
			{Ablation experiments and evaluations of the proposed GGHL on the DOTA dataset}}
		\vspace{-0.5em}
		\resizebox{\textwidth}{!}{\setlength{\tabcolsep}{1mm}{
				\begin{tabular}{c|c|c|c|c|c|c|c|c|c|c|ccccc}
					\hline\hline
					\multicolumn{2}{c|}{\multirow{3}{*}{Methods}}           & \begin{tabular}[c]{@{}c@{}}Data \\ Augmentation\end{tabular} & \multicolumn{3}{c|}{Label Assignment} & \multicolumn{3}{c|}{OBB Representation} & \multicolumn{2}{c|}{\begin{tabular}[c]{@{}c@{}}Objective \\ Function\end{tabular}} & \multirow{3}{*}{mAP} & \multirow{3}{*}{\begin{tabular}[c]{@{}c@{}}Inference\\ Speed\\ (fps)\end{tabular}} & \multirow{3}{*}{\begin{tabular}[c]{@{}c@{}}FLOPs\\ (G)\end{tabular}} & \multirow{3}{*}{\begin{tabular}[c]{@{}c@{}}Model \\ Parameters \\ (MB)\end{tabular}} & \multirow{3}{*}{\begin{tabular}[c]{@{}c@{}}Number of\\ Hyper-\\ parameters\end{tabular}} \\ \cline{3-11}
					\multicolumn{2}{c|}{}& MSC & Anchor-Box  & \begin{tabular}[c]{@{}c@{}}Standard \\ Gaussian\end{tabular} & OLA & \begin{tabular}[c]{@{}c@{}}Vanilla\\ Head\end{tabular} & ORC & \begin{tabular}[c]{@{}c@{}}ORC\\ -OWAM\end{tabular}  & \begin{tabular}[c]{@{}c@{}}Vanilla \\ Loss\end{tabular} & JOL& & & & & \\ \hline
					\multirow{2}{*}{\begin{tabular}[c]{@{}c@{}}Anchor\\-based\end{tabular}} & Vanilla-AB&& $\checkmark$ & & & $\checkmark$ &  &  &$\checkmark$ & & 72.55 & 38.77 & 130.93 & 68.87 & 19 \\
					& Vanilla-AB (MSC)&$\checkmark$& $\checkmark$ & & & $\checkmark$ &  & & $\checkmark$ & & 74.64 & 38.77 & 130.93 & 68.87 & 19  \\ \hline
					\multirow{4}{*}{\begin{tabular}[c]{@{}c@{}}Anchor\\-free\end{tabular}}  & Vanilla-AF (Baseline) &$\checkmark$& & $\checkmark$&  &  & $\checkmark$ & & $\checkmark$ & & 72.33 & 42.39 & 121.84 & 62.59 & 3\\
					& OLA + ORC&$\checkmark$& & &$\checkmark$& & $\checkmark$& & $\checkmark$ & & 74.89                & 42.39  & 121.84 & 62.59 & 3  \\
					& OLA + ORC-OWAM & $\checkmark$& & & $\checkmark$ & & & $\checkmark$&$\checkmark$& & 75.43 & 42.39 & 121.84 & 62.59 & 3  \\
					& GGHL &$\checkmark$& & &$\checkmark$ & & &$\checkmark$& & $\checkmark$& \textbf{76.95} & \textbf{42.39} & \textbf{121.84} & \textbf{62.59}  & \textbf{3} \\ 
					\hline\hline
		\end{tabular}}}}\vspace{0.5em}
\justifying{Note: Bold indicates the best result. The size of the input image is 800×800 pixels. The unit G is Giga, which represents 1×10$^{-9}$. The unit MB represents 1×10$^{-6}$ bytes. The inference speed only includes the network inference speed without pre-processing \& post-processing. Vanilla-AB represents the anchor-based Vanilla model, and Vanilla-AF represents the anchor-free Vanilla model.}
\end{table*}
\begin{table*}[htbp]
	\centering
	\vspace{-1em}
	\setlength{\tabcolsep}{1.25mm}{
		\caption{\label{table:3}
			{More detailed mAP (\%) results of ablation experiments on the DOTA dataset}}
		\vspace{-0.5em}
		\begin{threeparttable}
			\centering
			\renewcommand\arraystretch{1.15}
			\resizebox{\linewidth}{!}{
				\begin{tabularx}{\textwidth}{c|ccccccccccccccc|c}
					\hline\hline
					Methods & PL & BD & BR & GTF & SV & LV & SH & TC & BC & ST & SBF & RA & HA & SP & HC & mAP \\ \hline
					Vanilla-AB & 89.66 & 81.39 & 41.09 & 66.99 & 69.22 & 72.36 & \textbf{86.76} & \textbf{90.89} & 84.79 & 85.20 & 56.12 & 65.41 & 69.29 & 67.64 & 61.48 & 72.55 (Baseline)   \\
					Vanilla-AB (MSC) & 89.45 & 82.57 & 44.82 & 77.32 & 75.94 & 75.44 & 85.94 & 90.82 & 86.26 & 84.14 & 66.03 & 64.93 & 67.26 & 65.90 & 62.76 & 74.64 (+2.09) \\
					\hline
					Vanilla-AF (Baseline) & 89.12 & 82.23 & 39.10 & 75.16 & 70.97 & 74.45 & 86.03 & 90.85 & 85.98 & 84.11 & 57.47 & 58.43 & 66.08 & 66.32 & 58.73 & 72.33 (Baseline) \\
					OLA + ORC & 89.69 & 81.24 & 44.31 & \textbf{79.04} & 72.63 & 72.63 & 85.95 & 90.85 & 87.00 & 85.25 & 68.39 & 67.25 & 67.66 & 67.87 & 60.50 & 74.89 (+2.56) \\
					OLA + ORC-OWAM & 89.25 & 82.56 & 44.47 & 77.21 & 73.41 & 80.00 & 83.67 & 90.81 & 87.63 & 84.03 & \textbf{68.93} & 65.38 & 69.37 & 67.26 & 67.55 & 75.43 (+3.10) \\				
					GGHL & \textbf{89.74} & \textbf{85.63} & \textbf{44.50} & 77.48 & \textbf{76.72} & \textbf{80.45} & 86.16 & 90.83 & \textbf{88.18} & \textbf{86.25} & 67.07 & \textbf{69.40} & \textbf{73.38} & \textbf{68.45} & \textbf{70.14} & \textbf{76.95 (+4.62)} \\
					\hline\hline
			\end{tabularx}}\vspace{0.5em}
	\end{threeparttable}}
\justifying{Note: The size of the input image is 800×800 pixels. Bold indicates the best result. Vanilla-AB represents the anchor-based Vanilla model, and Vanilla-AF represents the anchor-free Vanilla model.}
\end{table*}

Second, when the external control factors like data augmentation strategy are consistent, the performance of anchor-based Vanilla-AB and anchor-free Vanilla-AF is compared. For this direct modification from anchor-based to anchor-free strategy, each layer of FPN changes from predicting anchor boxes of three scales to directly predicting the standard Gaussian candidates, of which the number of output feature maps becomes one-third. Although the computational complexity (FLOPs), model size (Model Parameters) and the number of hyperparameters have been reduced, and the detection speed has become faster, the mAP has been reduced by 2.31. On one hand, the performance is reduced due to the absence of the anchor prior and the reduction of model parameters. On the other hand, as analyzed above, the circular positive candidate defined by the standard Gaussian is not suitable for oriented objects, especially BR, SV and other objects with obvious directionality. For objects with approximately square OBBs, such as PL and BD, the performance loss is not obvious.
\begin{figure}[bp]
	\vspace{-1.5em}
	\centering
	\epsfig{width=0.48\textwidth,file=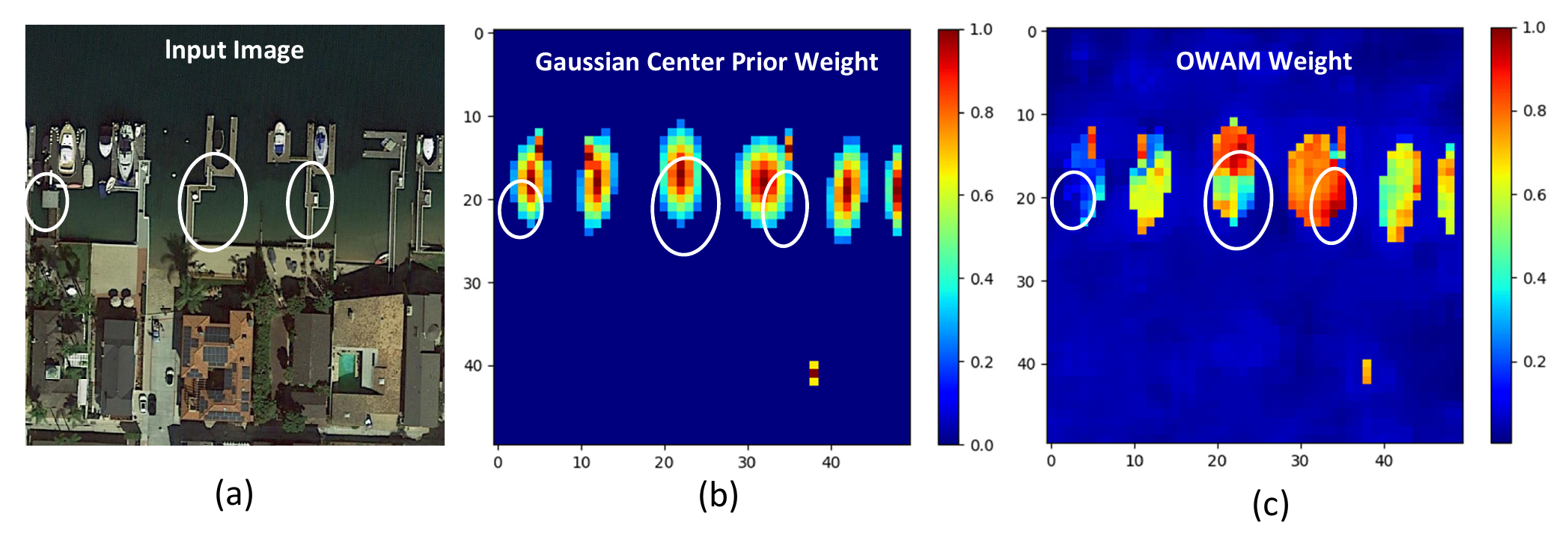}
	\vspace{-1em}
	\caption{The visualized feature maps of Gaussian center prior and learnable OBB regression confidence. (a) Input image. (b) Gaussian center prior. c) Learnable OBB regression confidence. Some typical non-Gaussian areas are marked with white circles.}\label{fig:8}
	\vspace{-0.5em}
\end{figure} 
\begin{figure}[bp]
	\centering
	\epsfig{width=0.48\textwidth,file=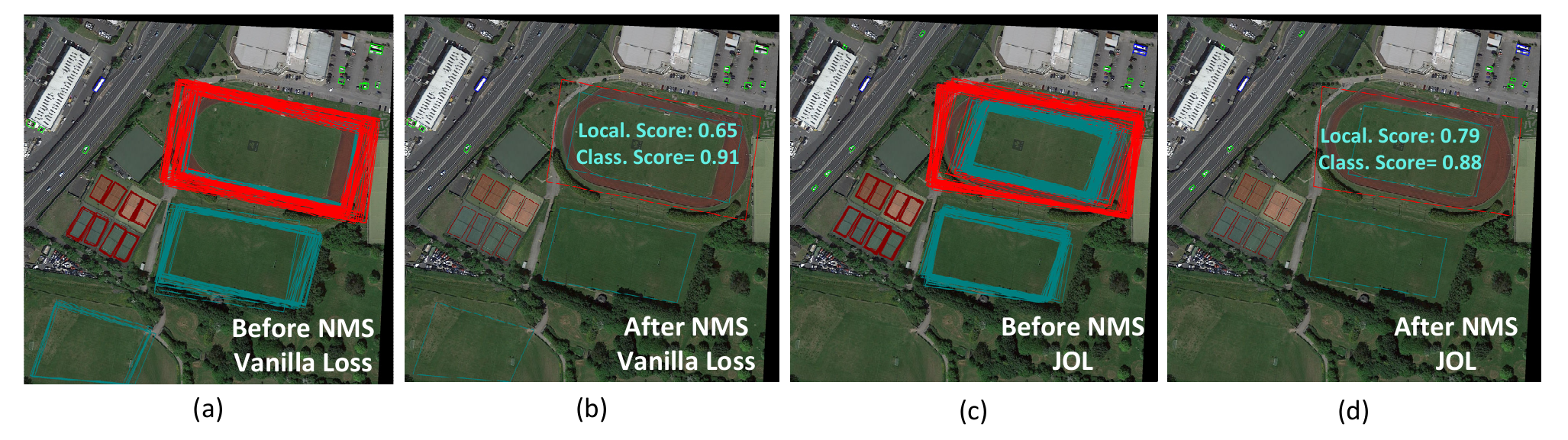}
	\vspace{-1em}
	\caption{A visual comparison of the results with and without JOL is on the validation dataset. a) Results before NMS using Vanilla Loss. b) Results after NMS using Vanilla Loss. c) Results before NMS using JOL. d) Results after NMS using JOL.}\label{fig:9}
	%\vspace{-0.5em}
\end{figure} 

\begin{table}[htbp]
	\centering
	\vspace{-2em}
	\setlength{\tabcolsep}{2.5mm}{
		\caption{\label{table:4}
			{Experiments with different values of $T_{IoU}$, $\tau$, $\vartheta$, and $\xi$ on the DOTA dataset}}
		\vspace{-1em}
		{\renewcommand\arraystretch{1.2}
			\begin{tabular}{cc|cc|cc|cc}
					\hline\hline
					$T_{IoU}$ & mAP & $\tau$  & mAP & $\vartheta$ & mAP & $\xi$ & mAP \\	
					\hline
					\multirow{2}{*}{0.3} & \multirow{2}{*}{\textbf{76.95}} & 2.0 & 74.83 & 0.1 & 75.42 & \multirow{2}{*}{without} & \multirow{2}{*}{76.13}\\
				    ~ & ~ & 2.5 & 75.71 & 0.3 & 76.67 & ~ & ~\\
					\multirow{2}{*}{0.4} & \multirow{2}{*}{76.18} & 3.0 & \textbf{76.95} & 0.5 & \textbf{76.95} & \multirow{2}{*}{with} & \multirow{2}{*}{\textbf{76.95}} \\
					~ & ~ & 3.5 & 74.99 & 1.0 & 74.89 & ~ & ~\\
					\hline\hline
			\end{tabular}}}\vspace{0.5em}
\justifying{Note: Bold indicates the best result. When evaluating one variable, the other variables are fixed to take the optimal value.}
 	\vspace{-1.8em}
\end{table}
%\multirow{2}{*}{GGHL}
Third, based on the baseline, i.e., Vanilla-AF, the proposed OLA and ORC are used to make the positive candidate region conform to the shape and direction characteristics of the objects. This improvement makes the mAP increase by 2.56. The object candidates of ORC are further improved by OWAM, i.e., ORC-OWAM, and mAP is further improved by 0.54. For the non-Gaussian center prior objects analyzed previously, like the harbor (HA), the performance improves more. The visualized feature maps of the CNN output layer in Fig.~\ref{fig:8} verify this claim. Further using the proposed JOL, the mAP increases by 1.52. The visualized results before and after NMS without/with JOL are shown in Fig.~\ref{fig:9}. When JOL is not used, the prediction result retained after NMS sorting may have a high classification score but a low location score, such as the soccer-ball field (SBF) in Fig.~\ref{fig:9} (b).  Note that the location of the SBF in Fig.~\ref{fig:9} (b) has a large deviation. Instead, when JOL is used, a more consistent result with the highest scores of OBB regression and classification is obtained, as shown in Fig.~\ref{fig:9} (d). The mutual promotion of the three components of GGHL is more significant, and the mAP has reached 76.95. It is an increase of 4.62 (6.39\%) compared to the baseline. Compared with the anchor-based method, i.e., Vanilla-AB (MSC), it increases by 2.31 (3.09\%), and has faster speed, lower computational complexity and model size, and saves the recessive cost of hyperparameter adjustments. In summary, the ablation experiments and visualization feature maps support the claims analyzed in the introduction and verify the effectiveness of each component from quantitative and qualitative perspectives. In addition, experiments are performed on different values of $T_{IoU}$, $\tau$, $\vartheta$, and $\xi$, the results are listed in Table~\ref{table:4}. When $T_{IoU}=0.3$, $\tau=3$ and $\vartheta=0.5$ with area normalization factor $\xi$, the proposed GGHL has the best performance on the DOTA dataset. Multi-scale assignment controlled by $\tau$ has a greater impact on model performances. Designing a scale assignment strategy for multi-scale objects is a direction worth continuing to study in the future. Using or not using OWAM has a greater impact on the results (mAP gap reaches 2.06), but the value of $\vartheta$ between 0.3-0.5 has little impact on mAP. Using area normalization significantly improves mAP, which confirms its effectiveness.

Furthermore, Table~\ref{table:4_A} analyzes the performance of ORC with/without refined approximation (RA) and Gliding Vertex \cite{xu2020gliding}. The ablation experiments demonstrate the effectiveness of RA and the effectiveness of the proposed label assignment strategy compared with the anchor-based strategy.
\begin{table}[tbp]
	\centering
	\vspace{-2em}
	\setlength{\tabcolsep}{5.7mm}{
		\caption{\label{table:4_A}
			{Ablation Experiment of ORC and RA on the DOTA dataset}}
		\vspace{-1em}
		{\renewcommand\arraystretch{1.1}
				\begin{tabular}{cccc}
					\hline\hline
					Methods & minRect & RA & mAP \\
					\hline
					\multirow{2}{*}{Gliding Vertex \cite{xu2020gliding}} & \checkmark & & 74.64\\
					~ & & \checkmark & 75.32\\
					\hline
					\multirow{2}{*}{GGHL} & \checkmark & & 76.28 \\
					~ & & \checkmark & \textbf{76.95}\\
					\hline\hline
	\end{tabular}}}\vspace{0.5em}
	\justifying{Note: Bold indicates the best result. "minRect" denotes the approximate method using minimum circumscribed rectangle.}
	\vspace{-1.5em}
\end{table}

\begin{table}[tbp]
	\centering
	\vspace{-1em}
	\renewcommand\arraystretch{1.3}
	\setlength{\tabcolsep}{1.3mm}{
		\caption{\label{table:5}
			{Ablation experiments and evaluations of the proposed GGHL on the DOTA dataset}}
		\vspace{-0.5em}
		\resizebox{0.49\textwidth}{!}{\setlength{\tabcolsep}{0.8mm}{
	\begin{threeparttable}
	\begin{tabular}{ccccc}
		\hline\hline
		\multicolumn{1}{c|}{{}}  & \multicolumn{1}{c|}{{}} & \multicolumn{1}{c|}{{}} & \multicolumn{1}{c|}{{}} & {}  \\
		\multicolumn{1}{c|}{\multirow{-2}{*}{{Methods}}}                                            & \multicolumn{1}{c|}{\multirow{-2}{*}{{Anchor}}} & \multicolumn{1}{c|}{\multirow{-2}{*}{{Backbone}}} & \multicolumn{1}{c|}{\multirow{-2}{*}{{mAP}}}                                      & \multirow{-2}{*}{{\begin{tabular}[c]{@{}c@{}}Inference\\ Speed (fps)\end{tabular}}}  \\ \hline
		\multicolumn{1}{c|}{{\begin{tabular}[c]{@{}c@{}}R-CenterNet \cite{zhou2019objects}\end{tabular}}} & \multicolumn{1}{c|}{{AF}} & \multicolumn{1}{c|}{{DarkNet53}}                  & \multicolumn{1}{c|}{{\begin{tabular}[c]{@{}c@{}}72.08\\ \end{tabular}}} & { \begin{tabular}[c]{@{}c@{}}46.31\\ \end{tabular}}                          \\
		\multicolumn{1}{c|}{{\begin{tabular}[c]{@{}c@{}}R-CenterNet \cite{zhou2019objects} + GGHL1\end{tabular}}}  & \multicolumn{1}{c|}{{AF}} & \multicolumn{1}{c|}{{DarkNet53}} & \multicolumn{1}{c|}{{73.63 (+1.57)}}  & {46.31 (+0)}  \\ \hline
		\multicolumn{1}{c|}{{\begin{tabular}[c]{@{}c@{}}R-FCOS-P5 \cite{tian2019fcos}\end{tabular}}} & \multicolumn{1}{c|}{{AF}} & \multicolumn{1}{c|}{{DarkNet53}} & \multicolumn{1}{c|}{{ \begin{tabular}[c]{@{}c@{}}73.48 \end{tabular}}} & {\begin{tabular}[c]{@{}c@{}}42.39\\ \end{tabular}}                          \\
		
		\multicolumn{1}{c|}{{\begin{tabular}[c]{@{}c@{}}R-FCOS-P5 + AutoAssign\cite{zhu2020autoassign}\end{tabular}}} & \multicolumn{1}{c|}{{AF}} & \multicolumn{1}{c|}{{DarkNet53}} & \multicolumn{1}{c|}{{ \begin{tabular}[c]{@{}c@{}}75.34 (+1.86) \end{tabular}}} & {\begin{tabular}[c]{@{}c@{}}42.39 (+0)\\ \end{tabular}}                          \\
		\multicolumn{1}{c|}{{\begin{tabular}[c]{@{}c@{}}R-FCOS-P5 \cite{tian2019fcos} + GGHL2\end{tabular}}}    & \multicolumn{1}{c|}{{AF}}  & \multicolumn{1}{c|}{{DarkNet53}} & \multicolumn{1}{c|}{{76.57 (+3.09)}} & {42.39 (+0)}  \\ \cline{1-5}
		\multicolumn{1}{c|}{{NPMMR-Det \cite{9364888}}} & \multicolumn{1}{c|}{{AB}} & \multicolumn{1}{c|}{{DarkNet53}} & \multicolumn{1}{c|}{{\begin{tabular}[c]{@{}c@{}}75.67\\ \end{tabular}}} & { \begin{tabular}[c]{@{}c@{}}32.52\\ \end{tabular}} \\
		\multicolumn{1}{c|}{{\begin{tabular}[c]{@{}c@{}}NPMMR-Det \cite{9364888} + GGHL\end{tabular}}} & \multicolumn{1}{c|}{{AF}}  & \multicolumn{1}{c|}{{DarkNet53}} & \multicolumn{1}{c|}{{77.74 (+2.07)}}& {35.98 (+3.46)} \\ \hline
		\multicolumn{1}{c|}{{LO-Det 608 \cite{huang2021lo}}}& \multicolumn{1}{c|}{{AB}}& \multicolumn{1}{c|}{{MobileNetv2}} & \multicolumn{1}{c|}{{\begin{tabular}[c]{@{}c@{}}66.17\\ \end{tabular}}} & { \begin{tabular}[c]{@{}c@{}}60.01\\ \end{tabular}}\\
		\multicolumn{1}{c|}{{\begin{tabular}[c]{@{}c@{}}LO-Det \cite{huang2021lo} + GGHL 608\end{tabular}}} & \multicolumn{1}{c|}{{AF}} & \multicolumn{1}{c|}{{MobileNetv2}} & \multicolumn{1}{c|}{{71.26 (+5.09)}} & {62.07 (+2.06)}\\
		\hline\hline
	\end{tabular}\vspace{0.5em}
\end{threeparttable}
}}
}
\justifying{Note: The size of the default input image is 800×800 pixels. For the lightweight detector LO-Det, the resolution of the CNN input layer is 608×608 pixels. AF represents anchor-free methods, and AB represents anchor-based methods. The inference speed only includes the network inference speed without post-processing. GGHL1: For embedding GGHL into R-CenterNet, OLA and ORC are used but only the center point is taken as a positive candidate like CenterNet; the original loss of CenterNet is still used but weighted and regularized. GGHL2: OLA and ORC are used, and loss is in the form of FCOS, but the Centerness is calculated by a two-dimensional Gaussian function.}
\vspace{-1.7em}
\end{table}
\textbf{2) Ablation experiments on different baseline models.} To further verify the effectiveness and versatility of the proposed GGHL, several state-of-the-art models are selected as the baselines for ablation experiments. The results of using GGHL on other models on the DOTA dataset are listed in Table~\ref{table:5}. First, the two popular anchor-free models, CenterNet \cite{zhou2019objects} and FCOS \cite{tian2019fcos}, are selected as baselines. Since these two models are designed for the ordinary OD task, they have been modified for the AOOD task. Among them, the modified CenterNet, i.e., R-CenterNet, uses Darknet53, which is simpler and the same as the Vanilla Model, instead of the original complex Hourglass-104 as the backbone. The modified FCOS, i.e., R-FCOS-P5, also uses the same backbone, using a 3-layer (P3-P5) FPN structure. The mAPs of the modified R-CenterNet and R-FCOS-P5 on the DOTA dataset are 72.08 and 73.48, respectively. The proposed GGHL is employed into these baselines to improve their label assignment strategy, and the mAP on the DOTA dataset is increased by 1.57 on R-CenterNet and 3.09 on R-FCOS-P5. Since these two baselines are anchor-free models originally, the model inference speed remains unchanged after the employment of GGHL. In addition, the performance of AutoAssign \cite{zhu2020autoassign}, which also uses an adaptive weighting strategy, has been tested based on R-FCOS \cite{tian2019fcos}. Although it also improves the performance of baseline, its performance on the AOOD task is inferior to the proposed GGHL because the Gaussian prior of AutoAssign \cite{zhu2020autoassign} is not directional and is shared with each category.

Second, NPMMR-Det \cite{9364888}, one of the latest methods in the typical AOOD task of remote sensing object detection, is selected as the baseline. It is an anchor-based model that balances accuracy and speed through CNN feature refinement design. The experiments indicate that using GGHL to improve it increases mAP by 2.07 and the speed by 3.46 fps. This result not only validates the effectiveness of GGHL for improving the anchor-based model, but also demonstrates that GGHL is also compatible with more complex CNN designs.

\begin{figure}[tbp]
	%\vspace{-1em}
	\centering
	\epsfig{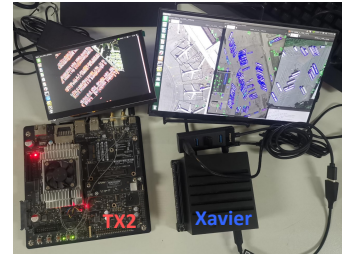}
	\vspace{-1em}
	\caption{The experiments on embedded devices.}\label{fig:10}
	\vspace{-1em}
\end{figure}

Third, the latest lightweight AOOD model LO-Det \cite{huang2021lo} is selected as the baseline to verify the effectiveness of the proposed GGHL on the lightweight model. The experimental results show that after the employment of GGHL improvements, LO-Det’s mAP on the DOTA dataset has increased by 5.09 (+7.69\%), and the speed has increased by 2.06 fps. Furthermore, experiments have also been carried out on embedded devices, and the results are shown in Table~\ref{table:6} and Fig.~\ref{fig:10}. The speed of improved LO-Det on TX2 and Xavier embedded devices has increased by 0.69 fps and 1.60 fps, respectively. FLOPs are reduced by 0.12 G, and model parameters are reduced by 0.21 MB. The improved performance verifies the application friendliness of the proposed GGHL for lightweight models on embedded devices.
\begin{table}[tbp]
	\centering
	\vspace{-1em}
	\renewcommand\arraystretch{1.3}
	\setlength{\tabcolsep}{1.3mm}{
		\caption{\label{table:6}
			{Ablation experiments and evaluations of the proposed GGHL on the DOTA dataset}}
		\vspace{-0.5em}
		\resizebox{0.49\textwidth}{!}{\setlength{\tabcolsep}{0.5mm}{
	\begin{threeparttable}
	\begin{tabular}{c|ccccccc}
		\hline\hline
		{}& {} & {} & {}  & {} & {} & { }& { } \\
		{}& {} & {} & {}  & {} & {} & { }& { } \\
		\multirow{-3}{*}{{Modules}} & \multirow{-3}{*}{{mAP}}& \multirow{-3}{*}{{\begin{tabular}[c]{@{}c@{}}Speed 1\\ (fps)\end{tabular}}} & \multirow{-3}{*}{{\begin{tabular}[c]{@{}c@{}}Speed 2\\ (fps)\end{tabular}}} & \multirow{-3}{*}{{\begin{tabular}[c]{@{}c@{}}Speed 3\\ (fps)\end{tabular}}} & \multirow{-3}{*}{{\begin{tabular}[c]{@{}c@{}}FLOPs\\ (G)\end{tabular}}}   & \multirow{-3}{*}{{\begin{tabular}[c]{@{}c@{}}Model \\ Parameters\\ (MB)\end{tabular}}} & \multirow{-3}{*}{{\begin{tabular}[c]{@{}c@{}}Number of\\ Hyper\\ -parameters\end{tabular}}} \\ \hline
		{LO-Det 608 \cite{huang2021lo}} & {66.17}& {60.01}  & {6.99}& {22.12} & {6.42}& {6.93} & {19}\\
		\hline
		{}& {} & {} & {} & {} & {} & {} & {}\\
		\multirow{-2}{*}{{\begin{tabular}[c]{@{}c@{}}LO-Det \cite{huang2021lo}\\ + GGHL 608\end{tabular}}} & \multirow{-2}{*}{{\begin{tabular}[c]{@{}c@{}}71.26\\ (+5.09)\end{tabular}}} & \multirow{-2}{*}{{\begin{tabular}[c]{@{}c@{}}62.07\\ (+2.06)\end{tabular}}} & \multirow{-2}{*}{{\begin{tabular}[c]{@{}c@{}}7.68\\ (+0.69)\end{tabular}}}  & \multirow{-2}{*}{{\begin{tabular}[c]{@{}c@{}}23.72\\ (+1.60)\end{tabular}}} & \multirow{-2}{*}{{\begin{tabular}[c]{@{}c@{}}6.30\\ (-0.12)\end{tabular}}} & \multirow{-2}{*}{{\begin{tabular}[c]{@{}c@{}}6.72\\ (-0.21)\end{tabular}}} & \multirow{-2}{*}{{\begin{tabular}[c]{@{}c@{}}3\\ (-16)\end{tabular}}}   \\ \hline\hline
	\end{tabular}\vspace{0.5em}
\end{threeparttable}}}
}
\justifying{Note: The unit G is Giga, which represents 1×10$^{-9}$. The unit MB represents 1×10$^{-6}$ bytes. Speed 1 is the speed on RTX 3090 GPU, speed 2 is the speed on NVIDIA Jetson TX2, Speed 3 is the speed on NVIDIA Jetson AGX Xavier. The inference speed (average of 10 tests) only includes the network inference speed without post-processing.}
\vspace{-1em}
\end{table}

\vspace{-0.5em}
\subsection{Comparative Experiments and Analysis}
Comparative experiments are conducted extensively on several public AOOD datasets from different typical scenarios to compare the performance of the proposed GGHL and the state-of-the-art methods.

\begin{figure}[htb]
	\vspace{-0.5em}
	\centering
	\epsfig{width=0.48\textwidth,file=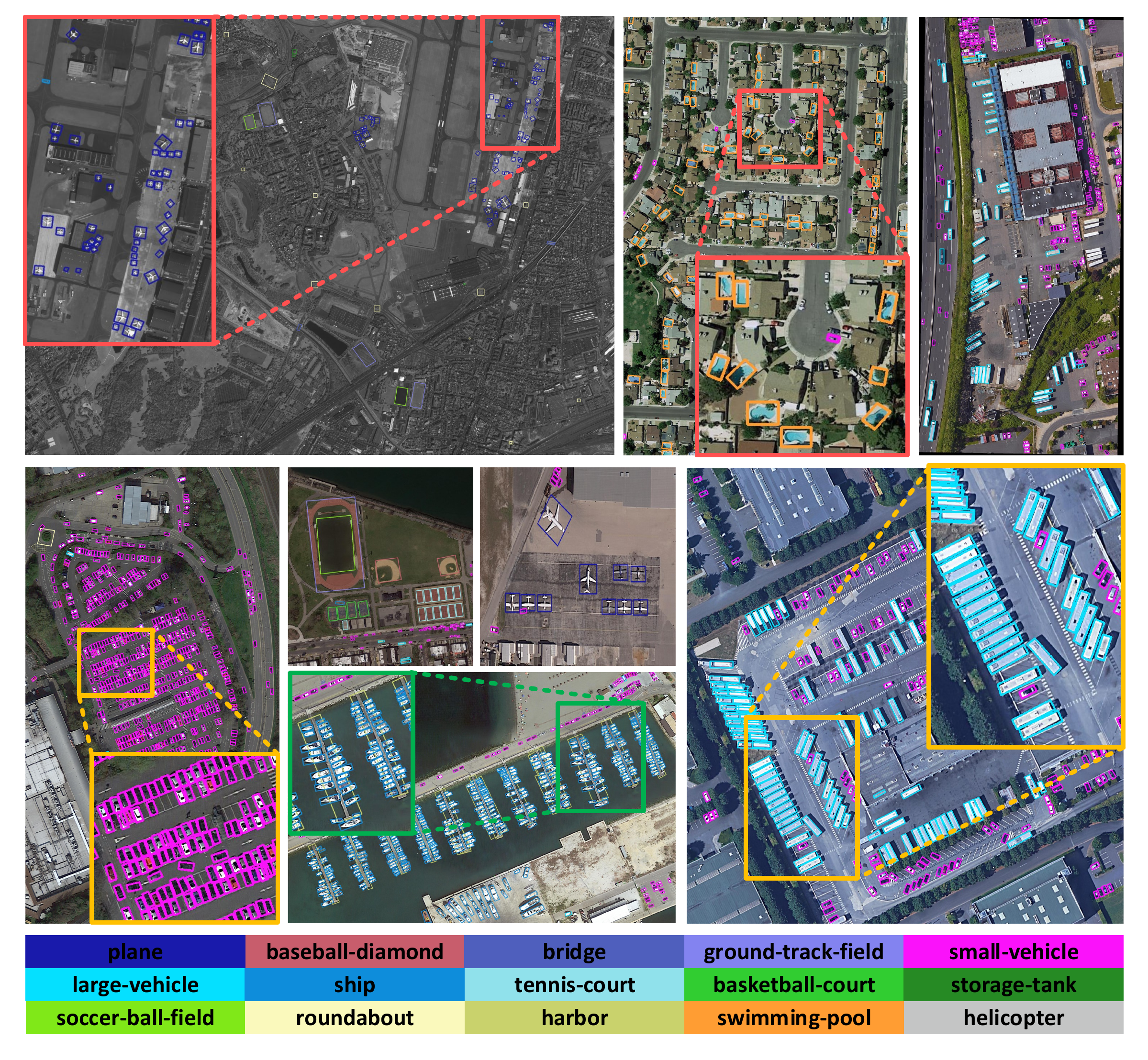}
	\vspace{-1em}
	\caption{Visualization Results of the proposed GGHL on the DOTA Dataset.}\label{fig:11}
	\vspace{-1em}
\end{figure}

\textbf{1) Comparative Experiments on the DOTA dataset.}
\begin{table*}[!htb]
	\centering
	\vspace{-1em}
	\renewcommand\arraystretch{1.05}
	\setlength{\tabcolsep}{1mm}{
		\caption{\label{table:7}
			{Comparative experiments on the DOTA dataset}}
		\vspace{-0.5em}
		\resizebox{\textwidth}{!}{\setlength{\tabcolsep}{0.4mm}{
	\begin{threeparttable}
		\centering
		\resizebox{\textwidth}{!}{
			\begin{tabularx}{\textwidth}{c|c|c|c|ccccccccccccccc|c|c}
				\hline\hline
			\multirow{2}{*}{Methods}                                                            & \multirow{2}{*}{Year} & \multirow{2}{*}{Backbone} & \multirow{2}{*}{Anchor} & \multirow{2}{*}{PL} & \multirow{2}{*}{BD} & \multirow{2}{*}{BR} & \multirow{2}{*}{GTF} & \multirow{2}{*}{SV} & \multirow{2}{*}{LV} & \multirow{2}{*}{SH} & \multirow{2}{*}{TC} & \multirow{2}{*}{BC} & \multirow{2}{*}{ST} & \multirow{2}{*}{SBF} & \multirow{2}{*}{RA} & \multirow{2}{*}{HA} & \multirow{2}{*}{SP} & \multirow{2}{*}{HC} & \multirow{2}{*}{mAP} & \multirow{2}{*}{\begin{tabular}[c]{@{}c@{}}Speed\\ (fps)\end{tabular}} \\
			& &  &  &  &  &  &  & &   &  &  & &  &  & & & & &  &  \\ \hline
			ROI Trans. \cite{ding2019learning}  & 2019  & ResNet101 & AB & 88.53& 77.91 & 37.63& 74.08 & 66.53& 62.97  & 66.57& 90.50& 79.46& 76.75& 59.04  & 56.73& 62.54  & 61.29 & 55.56& 67.74& 7.80 \\
			RSDet \cite{qian2019learning}  & 2019 & ResNet101& AB & 89.80 & 82.90  & 48.60& 65.20& 69.50 & 70.10 & 70.20 & 90.50  & 85.60 & 83.40 & 62.50 & 63.90& 65.60 & 67.20& 68.00  & 72.20 & -  \\
			SCRDet \cite{yangSCRDetMoreRobust2019} & 2019  & ResNet101  & AB & 89.98& 80.65 & 52.09& 68.36 & 68.36 & 60.32 & 72.41 & 90.85& 87.94& 86.86 & 65.02 & 66.68  & 66.25  & 68.24  & 65.21  & 72.61  & 9.51\\
			R$^3$Det \cite{yang2019r3det} & 2019 & ResNet152 & AB & 89.80 & 83.77 & 48.11 & 66.77 & 78.76 & 83.27 & 87.84 & 90.82& 85.38 & 85.51& 65.67 & 62.68 & 67.53 & \textbf{78.56}& 72.62 & 76.47  & 10.53 \\
			Gliding Vertex \cite{xu2020gliding} & 2019 & ResNet101 & AB & 89.64 & 85.00 & 52.26 & 77.34 & 73.01 & 73.14 & 86.82 & 90.74& 79.02  & 86.81 & 59.55  & 70.91 & 72.94 & 70.86 & 57.32 & 75.02 & 13.10\\
			O$^2$-DNet \cite{wei2020oriented} & 2020  & Hourglass-104 & AF & 89.31 & 82.14  & 47.33  & 61.21  & 71.32 & 74.03 & 78.62& 90.76 & 82.23 & 81.36  & 60.93 & 60.17 & 58.21 & 66.98 & 61.03 & 71.04 & - \\
			BBAVectors \cite{yi2020oriented} & 2020 & ResNet101& AF & 88.35& 79.96& 50.69 & 62.18& 78.43& 78.98& 87.94 & 90.85 & 83.58& 84.35& 54.13& 60.24 & 65.22 & 64.28 & 55.70 & 72.32 & 18.37 \\
			DRN \cite{pan2020dynamic} & 2020  & Hourglass-104 & AF & 89.71 & 82.34 & 47.22 & 64.10 & 76.22 & 74.43 & 85.84 & 90.57 & 86.18& 84.89& 57.65& 61.93 & 69.30 & 69.63 & 58.48 & 73.23 & - \\
			CSL \cite{yang2020arbitrary}& 2020 & ResNet152 & AB & \textbf{90.25}  & 85.53 & 54.64  & 75.31 & 70.44  & 73.51 & 77.62 & 90.84 & 86.15 & 86.69  & 69.60 & 68.04 & 73.83 & 71.10 & 68.93 & 76.17 & - \\
			S$^2$A-Net \cite{han2021align}& 2020& ResNet50  & AB  & 89.07  & 82.22 & 53.63  & 69.88 & \textbf{80.94}  & 82.12   & 88.72  & 90.73 & 83.77 & 86.92 & 63.78 & 67.86 & 76.51 & 73.03& 56.60 & 76.38 & 17.60\\
			S$^2$A-Net \cite{han2021align}& 2020 & ResNet101 & AB & 88.89 & 83.60 & \textbf{57.74} & \textbf{81.95} & 79.94 & 83.19  & 89.11& 90.78& 84.87 & \textbf{87.81} & 70.30& 68.25 & \textbf{78.30}& 77.01 & 69.58 & \textbf{79.42} & 13.79\\
			CFC-Net \cite{ming2021cfc}& 2021 & ResNet50  & AB & 89.08 & 80.41 & 52.41 & 70.02 & 76.28 & 78.11 & 87.21 & \textbf{90.89}      & 84.47  & 85.64 & 60.51 & 61.52 & 67.82 & 68.02 & 50.09 & 73.50 & 17.81 \\
			RIDet-O (RIL) \cite{ming2021optimization}& 2021 & ResNet101 & AB & 88.94 & 78.45 & 46.87 & 72.63 & 77.63 & 80.68 & 88.18 & 90.55 & 81.33 & 83.61 & 64.85 & 63.72 & 73.09 & 73.13 & 56.87 & 74.70 & 13.36\\
			S$^2$A-Net + RIL \cite{ming2021optimization}& 2021  & ResNet50 & AB & 89.31 & 80.77 & 54.07 & 76.38 & 79.81 & 81.99 & \textbf{89.13}& 90.72 & 83.58& 87.22 & 64.42 & 67.56 & 78.08 & 79.17 & 62.07 & 77.62 & 17.25\\
			\begin{tabular}[c]{@{}c@{}}RetinaNet-GWD \cite{yang2021rethinking}\end{tabular} & 2021 & ResNet152 & AB  & 86.14  & 81.59 & 55.33& 75.57& 74.20 & 67.34 & 81.75 & 87.48 & 82.80  & 85.46& 69.47  & 67.20 & 70.97 & 70.91 & \textbf{74.07} & 75.35 & 11.65                                                                  \\
			\begin{tabular}[c]{@{}c@{}}R$^3$Det-GWD \cite{yang2021rethinking}\end{tabular} & 2021 & ResNet50  & AB & 88.89  & 83.58 & 55.54 & 80.46 & 76.86 & 83.07 & 86.85 & 89.09 & 83.09 & 86.17 & \textbf{71.38} & 64.93 & 76.21 & 73.23 & 64.39 & 77.58 & 16.22 \\
			\begin{tabular}[c]{@{}c@{}}R$^3$Det-GWD \cite{yang2021rethinking}\end{tabular} & 2021 & ResNet152  & AB  & 88.99 & 82.26 & 56.62 & 81.40  & 77.04 & \textbf{83.90}& 86.56  & 88.97  & 83.63 & 86.48 & 70.45 & 65.58 & 76.41 & 77.30 & 69.21   & 78.32 & 10.50  \\ \hline
			\textbf{GGHL} & 2021& DarkNet53  & AF & 89.74& 85.63  & 44.50  & 77.48 & 76.72  & 80.45 & 86.16 & 90.83 & \textbf{88.18} & 86.25 & 67.07 & 69.40 & 73.38 & 68.45 & 70.14 & 76.95 & 42.30\\
			\textbf{NPMMR-Det-GGHL} & 2021  & DarkNet53 & AF  & 89.16 & \textbf{85.71} & 48.18  & 78.86 & 77.29  & 82.26  & 87.58& 90.88 & 88.04 & 86.86 & 65.74  & \textbf{69.82}& 74.44  & 70.75 & 70.47 & 77.74 & 35.98  \\
			\textbf{LO-Det-GGHL}  & 2021 & MobileNetv2 & AF & 89.66 & 83.02 & 38.55 & 77.09  & 72.57  & 71.86  & 82.47  & 90.78  & 78.05   & 83.56 & 47.74 & 67.83  & 64.21  & 67.83  & 54.16 & 71.26  & \textbf{62.07} \\
		\hline\hline
		\end{tabularx}}\vspace{0.5em}
	\end{threeparttable}}}}
\justifying{Note: Bold font indicates the best results. AF represents anchor-free methods, and AB represents anchor-based methods. The inference speed only includes the network inference speed (batch size=1) without post-processing on an RTX 3090 GPU. When testing other methods, their open source codes are used. Since the deep learning frameworks are different, there may be slight relative errors in the test speed. Some methods’ codes are not open-source, which is indicated by “-”. Regarding some methods, we have tried our best but failed to reproduce the results shown in their original papers, so the best results reported by them are shown in the Table. To align the other tricks of GGHL and GWD, the result of combining tricks for GWD (DA+MS+MSC) is selected as a comparison.} 
\vspace{-1.5em}
\end{table*}%\cite{redmon2017yolo9000}
In the AOOD task, most methods use the DOTA aerial remote sensing data \cite{qian2019learning,yang2019r3det,han2021align,ming2021cfc}, for performance comparison and analysis. In ablation experiments, this data set has been used to evaluate in detail the effectiveness, versatility, and performance of each component of the proposed GGHL. Table~\ref{table:7} provides comparison of detection performance for each category and some detail information of experimental implementation is supplied at its bottom. It can be observed that the detection accuracy of the proposed GGHL method (mAP=76.95) surpasses most of the AOOD methods in the past three years and has a very fast detection speed (fps=42.39). And GGHL has a new improvement in accuracy or speed by combining with AOOD methods, such as NPMMR-Det or LO-Det. Although GGHL's mAP is slightly lower than that of the excellent AOOD methods, such as S$^{2}$A-Net \cite{han2021align} and R$^{3}$Det-GWD \cite{yang2021rethinking} using larger backbones like ResNet101 or ResNet152, it runs faster than them. Moreover, GGHL is an anchor-free method with the lower recessive cost, which does not need to set and adjust prior hyperparameters. The visualization results of GGHL on the DOTA dataset including optical RGB images and panchromatic images are shown in Fig.~\ref{fig:11}. It can also be observed that the proposed GGHL accurately detects densely arranged objects benefited from the definition of positive locations and the label assignment strategy that more fit the objects’ shape and direction. Furthermore, comparative experiments are also conducted on the new versions of the DOTA datasets, i.e., DOTAv1.5 and DOTAv2.0 \cite{ding2021object}. The proposed GGHL is compared with the methods in the official latest benchmark \cite{ding2021object}, and the results are listed in Table ~\ref{table:7_A}, which indicates that the proposed GGHL not only achieves state-of-the-art performance in mAP but also has a very fast detection speed.
\begin{table}[tb]
	\centering
	\vspace{-0.5em}
	\renewcommand\arraystretch{1.3}
	\setlength{\tabcolsep}{1.3mm}{
		\caption{\label{table:7_A}
			{Comparative experiments on DOTAv1.0, DOTAv1.5, and DOTAv2.0  \cite{ding2021object} datasets}}
		\vspace{-1.em}
		\resizebox{0.49\textwidth}{!}{\setlength{\tabcolsep}{0.5mm}{
				\begin{threeparttable}	
					\begin{tabular}{c|ccc|c}
						\hline\hline
						Methods   & mAP@v1.0 & mAP@v1.5 & mAP@v2.0 & Speed (fps) \\ \hline
						RetinaNet OBB \cite{linFocalLossDense2017} & 66.28 & 59.16 & 46.68  & 12.10  \\
						Mask R-CNN \cite{ding2021object} & 70.71 & 62.67 & 49.47  & 9.70  \\
						Cascade Mask R-CNN \cite{ding2021object} & 70.96 & 63.41 & 50.04 & 7.20  \\
						Hybrid Task Mask \cite{ding2021object} & 71.21 & 63.40 & 50.34 & 7.90  \\
						Faster R-CNN OBB \cite{renFasterRCNNRealTime2017a} & 69.36 & 62.00 & 47.31 & 14.10  \\
						Faster R-CNN OBB + Dpool \cite{ding2021object} & 70.14 & 62.20 & 48.77 & 12.10  \\		
						Faster R-CNN H-OBB \cite{ding2021object} & 70.11 & 62.57 & 48.90 & 13.70  \\	
						Faster R-CNN OBB + RT \cite{ding2021object} & 73.76 & 65.03 & 52.81 & 12.40  \\						
						\hline
						\textbf{GGHL} & \textbf{73.98}  & \textbf{68.92} & \textbf{57.17} & \textbf{41.07}    \\ \hline\hline
					\end{tabular}\vspace{0.5em}
		\end{threeparttable}}}
	}
	\justifying{Note: Bold font indicates the best results. In order to make a fair comparison with the methods in the DOTAv2.0 benchmark \cite{ding2021object}, the experiments above do not use data augmentation and other tricks like these comparison methods. mAP@v1.0 denotes the results on the DOTAv1.0 dataset, mAP@v1.5 denotes the results on the DOTAv1.5 dataset, and mAP@v2.0 denotes the results on the DOTAv2.0 \cite{ding2021object} dataset. The speed of all methods are tested on a single NVIDIA Tesla V100 GPU.}
	\vspace{-1.5em}
\end{table}

\begin{figure}[htb]
	\vspace{-0.5em}
	\centering
	\epsfig{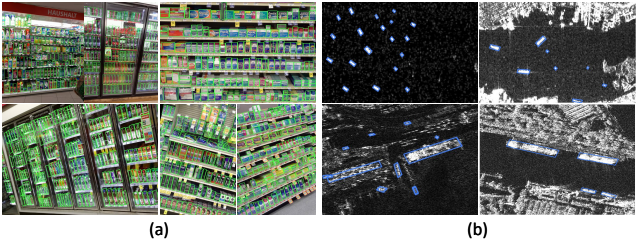}
	\vspace{-1em}
	\caption{Visualization Results of the proposed GGHL on (a) the SKU-110R Dataset, and (b) the SSDD+ Dataset.}\label{fig:12}
	\vspace{-1.5em}
\end{figure}

\begin{table}[t]
	\centering
	\vspace{-0.5em}
	\renewcommand\arraystretch{1.2}
	\setlength{\tabcolsep}{1.3mm}{
		\caption{\label{table:8}
			{Comparative experiments and evaluations of the proposed GGHL on the SKU-110R dataset}}
		\vspace{-0.5em}
		\resizebox{0.49\textwidth}{!}{\setlength{\tabcolsep}{0.5mm}{
				\begin{threeparttable}
					\begin{tabular}{c|c|c|c|c}
						\hline\hline
						Methods   & Backbone & Anchor & AP@0.75 & Speed (fps) \\ \hline
						YOLOv3-R \cite{pan2020dynamic} & DarkNet53& AB& 51.10  & 44.07\#  \\
						CenterNet-R \cite{pan2020dynamic}  & Hourglass-104 & AF & 61.10  & - \\
						DRN \cite{pan2020dynamic}  & Hourglass-104 & AF & 63.10  & - \\ \hline
						Vanilla-AF (Baseline) & DarkNet53  & AF  & 60.61 & 44.13  \\
						GGHL & DarkNet53  & AF & 63.73 & 44.13    \\ \hline\hline
					\end{tabular}\vspace{0.5em}
		\end{threeparttable}}}
	}
	\justifying{Note: AF represents anchor-free methods, and AB represents anchor-based methods. Some methods’ codes are not open-source, the unreported results of which is indicated by “-”. “\#” indicates the results we reproduced.}
	\vspace{-0.5em}
\end{table}
\begin{table}[t]
	\centering
	\vspace{-0.5em}
	\renewcommand\arraystretch{1.2}
	\setlength{\tabcolsep}{1.2mm}{
		\caption{\label{table:9}
			{Comparative experiments and evaluations of the proposed GGHL on the SSDD+ dataset}}
		\vspace{-0.5em}
		\resizebox{0.49\textwidth}{!}{\setlength{\tabcolsep}{0.5mm}{
				\begin{threeparttable}
					\begin{tabular}{c|c|c|c|c|c}
						\hline\hline
						\multirow{2}{*}{Methods} & \multirow{2}{*}{Backbone} & \multirow{2}{*}{Anchor} & \multirow{2}{*}{AP@0.3} & \multirow{2}{*}{AP@0.5} & \multirow{2}{*}{\begin{tabular}[c]{@{}c@{}}Speed\\ (fps)\end{tabular}} \\
						& &  &   &   & \\ \hline
						DRBox-v1 \cite{an2019drbox}  & VGG16   & AB  & 86.41  & - & -\\
						SDOE \cite{wang2018simultaneous}  & VGG16   & AB  & -  & 82.40   & -   \\
						DRBox-v2 \cite{an2019drbox} & VGG16  & AB  & 92.81 & 85.17\# & 49.09\#\\
						\hline
						Vanilla-AF (Baseline)& DarkNet53 & AF  & 95.09 & 87.04  & 43.87\\
						GGHL & DarkNet53  & AF  & 95.10& 90.22& 43.87  \\
						LO-Det + GGHL 608 & MobileNetv2  & AF   & 94.18 & 85.90 & 62.66 \\ 
						\hline\hline
					\end{tabular}\vspace{0.5em}
		\end{threeparttable}}}
	}
	\justifying{Note: AF represents anchor-free methods, and AB represents anchor-based methods. Some methods’ codes are not open-source, the unreported results of which is indicated by “-”. “\#” indicates the results we reproduced.}
	\vspace{-1.5em}
\end{table}
\textbf{2) Comparative Experiments on other AOOD datasets.} Further comparative experiments are carried out on multiple AOOD datasets including SUK110R \cite{pan2020dynamic} and SSDD+ \cite{li2017ship,an2019drbox,wang2018simultaneous} to verify the effectiveness of the proposed GGHL comprehensively. Their image types contain optical RGB image and polarized SAR image. The challenges they face include dense instances, noise interference, and diverse object appearances. The experimental results are shown in Table~\ref{table:8}-~\ref{table:9} and Fig.~\ref{fig:12}. On the SKU-110R dataset, the mAP and speed of the proposed GGHL surpass those of the existing methods. Compared with the baseline, GGHL makes the AP@0.75 increase by 1.3. On the SSDD+ dataset, compared with the baseline, GGHL makes the AP@0.5 increase by 3.18 without reducing the speed. The lightweight model LO-Det+GGHL not only has a slightly higher accuracy than DRBox-v2, but also has a faster speed. In summary, extensive and in-depth experiments on multiple datasets have verified the effectiveness of the proposed GGHL and demonstrated the evaluation results of its performance.

\section{Conclusions}
In this paper, a novel AOOD method, i.e., GGHL, was proposed. In GGHL, an anchor-free Gaussian OLA strategy reflecting objects’ shape and direction was designed to define and assign the positive candidate locations. An ORC mechanism was developed to indicate OBBs and an OWAM was presented to adjust the Gaussian center prior sample space for fitting the characteristics of different objects adaptively through the CNN learning. For refining the misalign optimal results of different subtasks in the constructed sample space during training, a JOL with area normalization and dynamic weighting was designed.

The extensive experiments on several public datasets have demonstrated the following: 1) The proposed GGHL has achieved state-of-the-art performance both in accuracy and speed on the AOOD task. The effectiveness of each component has been verified, and the claims made for each component are consistent and verified. 2) The proposed GGHL is a general framework that can be used to improve other AOOD methods in different scenarios. It improves the accuracy without reducing the detection speed, and does not require many anchor hyperparameters. 3) The proposed GGHL is friendly to the landing of CNN-based AOOD application, which improves the performance of lightweight AOOD models on embedded devices and saves a lot of hidden costs of tuning parameters. 

Despite the demonstrated benefits, the strategy of assigning labels to different scales adaptively and the abandonment of NMS to construct an end-to-end CNN model are still to be studied in the future.

The codes are available at https://github.com/Shank2358.

\section{Appendixes}
\normalsize
\subsection{The PDF of CNN in GGHL}
Since the above model is too complicated, let's start with the basic neuron model in a neural network to explain probability density function (PDF).

Without loss of generality, we may model the CNN model as follows:
\begin{equation}
\boldsymbol{\hat y} = nn\left( {\boldsymbol{x},\boldsymbol{\theta}} \right) + \boldsymbol{b},
\tag{A-1}
\label{eq:A-1}
\end{equation}
where $nn\left(  \cdot  \right)$ is a deterministic function related to the CNN; $\boldsymbol{x}$ is the input; $\boldsymbol{\hat y}$ denotes the output, i.e., the predictions of CNN; $\boldsymbol{\theta}$ denotes the vector composed of learnable parameters; $\boldsymbol{b}$ represents the bias vector, which is usually set to a zero vector in CNN. To simplify the derivation, this setting is also used here, i.e., $\boldsymbol{\hat y} = nn\left( {\boldsymbol{x},\boldsymbol{\theta}} \right)$. 

\textbf{1) PDF and loss function of OBB regression.} Define the ground truth of the prediction as $\boldsymbol{y}$. Define the error between the actual value and the predicted value as $\boldsymbol{\varepsilon}  = \boldsymbol{y} - \boldsymbol{\hat y}$, which is assumed to obey an i.i.d. Gaussian distribution with a mean of 0 and variance ${\sigma ^2}$. Therefore, the PDF is
\begin{equation}
	p\left( {\boldsymbol{y}\left| \boldsymbol{x} \right.;\boldsymbol{\theta} } \right) = p\left( \boldsymbol{\varepsilon}  \right) = \frac{1}{{\sigma \sqrt {2\pi } }}{e^{ - \frac{{{{\left( {\boldsymbol{y} - \boldsymbol{\hat y}} \right)}^2}}}{{2{\sigma ^2}}}}},
	\tag{A-2}
	\label{eq:A-2}
\end{equation}
which represents the probability density of $\boldsymbol{y}$ when $\boldsymbol{x}$ and $\boldsymbol{\theta}$ are given. Then, for multiple ${\boldsymbol{y}^{\left( i \right)}},{\rm{ }}i = 1,2, \cdots ,m$, in different locations of output layers, their joint PDF is
\begin{equation}
	\begin{array}{l}
		p\left( {{\boldsymbol{y}^{\left( 1 \right)}} \cdots {\boldsymbol{y}^{\left( m \right)}}\left| {{\boldsymbol{x}^{\left( 1 \right)}} \cdots {\boldsymbol{x}^{\left( m \right)}}} \right.;\boldsymbol{\theta} } \right)\\
		= \prod\limits_{i = 1}^m {p\left( {{\boldsymbol{y}^{\left( i \right)}}\left| {{\boldsymbol{x}^{\left( i \right)}}} \right.;\boldsymbol{\theta} } \right)}  = \prod\limits_{i = 1}^m {\frac{1}{{\sigma \sqrt {2\pi } }}{e^{ - \frac{{{{\left( {{\boldsymbol{y}^{\left( i \right)}} - {{\boldsymbol{\hat y}}^{\left( i \right)}}} \right)}^2}}}{{2{\sigma ^2}}}}}.}
	\end{array}
	\tag{A-3}
	\label{eq:A-3}
\end{equation}
Then, their joint likelihood function (LF) for $\boldsymbol{\theta}$ is
\begin{equation}
	\begin{array}{l}
		L\left( \boldsymbol{\theta}  \right) = \log p\left( {{\boldsymbol{y}^{\left( 1 \right)}} \cdots {\boldsymbol{y}^{\left( m \right)}}\left| {{\boldsymbol{x}^{\left( 1 \right)}} \cdots {\boldsymbol{x}^{\left( m \right)}}} \right.;\boldsymbol{\theta} } \right)\\
		= m\log \frac{1}{{\sigma \sqrt {2\pi } }} - \frac{1}{{2{\sigma ^2}}}\sum\limits_{i = 1}^m {{{\left( {{\boldsymbol{y}^{\left( i \right)}} - {{\boldsymbol{\hat y}}^{\left( i \right)}}} \right)}^2}}.
	\end{array}
	\tag{A-4}
	\label{eq:A-4}
\end{equation}

Now let us reconsider the process of using CNN to predict the shape of OBBs, which is a regression. Since the error of the OBB regression is assumed to obey an i.i.d. Gaussian distribution with a mean of 0 and variance ${\sigma ^2}$, the PDF of $\boldsymbol{obb}{_{x,y,m}} = \left[{{\boldsymbol{l}_{x,y,m}},{\boldsymbol{s}_{x,y,m}},ar{_{x,y,m}}} \right]$, when $\boldsymbol{x}{^{obb}_{x,y,m}}$ and $\boldsymbol{\theta}{^{obb}_{x,y,m}}$ are given, is
\begin{equation}
\begin{array}{l}
	p\left( {\boldsymbol{obb}{_{x,y,m}}\left| {obj{_{x,y,m}};\boldsymbol{x}_{x,y,m}^{obb}} \right.;\boldsymbol{\theta} _{x,y,m}^{obb}} \right)\\
	= \displaystyle \frac{1}{{\sigma \sqrt {2\pi } }}{\displaystyle e^{ - \frac{{{{\left( {\boldsymbol{obb}{_{x,y,m}} - {{\boldsymbol{\widehat {obb}}}_{x,y,m}}} \right)}^2}}}{{2{\sigma ^2}}}}}.
\end{array}
	\tag{A-5}
	\label{eq:A-5}
\end{equation}
Note that the prediction of OBB is performed under the condition of determined positive and negative locations, so the $obj{_{x,y,m}}$ is also one of the conditions in Eq.~\ref{eq:A-5}. The LF of parameters $\boldsymbol{\theta} _{x,y,m}^{obb}$, is
\begin{equation}
	\begin{array}{l}
L\left( {\boldsymbol{\theta}_{x,y,m}^{obb}} \right) =  - {\left( {\boldsymbol{obb}{_{x,y,m}} - {{\boldsymbol{\widehat {obb}}}_{x,y,m}}} \right)^2}.
	\end{array}
	\tag{A-6}
	\label{eq:A-6}
\end{equation}
According to MLE, the loss function at location ${\left( {x,y} \right)_m}$ is
\begin{equation}
\begin{array}{l}
	Loss\left( { \boldsymbol{obb}{_{x,y,m}}  - { \boldsymbol{\widehat {obb}}}_{x,y,m}} \right)  = \\ 
	\sum\limits_{k = 1}^4 {{{\left( {l_{x,y,m}^{\left( k \right)} - \hat l_{x,y,m}^{\!\left( k \right)}} \!\right)} ^2}} + \sum\limits_{k = 1}^4 {{{\left( {s_{x,y,m}^{\left( k \right)} - \hat s_{x,y,m}^{\left( k \right)}} \right)}^2}}  \\ + {\left( {a{r_{x,y,m}} - {{\widehat {ar}}_{x,y,m}}} \right)^2},
\end{array}
	\tag{A-7}
	\label{eq:A-7}
\end{equation}
where $\hat l_{x,y,m}^{\left( k \right)}$ is the $k$th component of $1 \times 4$-dimensional vector ${\boldsymbol{\hat l}_{x,y,m}}$, and $l_{x,y,m}^{\left( k \right)}$ is the $k$th component of $1 \times 4$-dimensional vector ${\boldsymbol{l}_{x,y,m}}$. $\hat s_{x,y,m}^{\left( k \right)}$ is the $k$th component of $1 \times 4$-dimensional vector ${\boldsymbol{\hat s}_{x,y,m}}$, and $s_{x,y,m}^{\left( k \right)}$ is the $k$th component of  $1 \times 4$-dimensional vector ${\boldsymbol{s}_{x,y,m}}$. Literature \cite{rezatofighiGeneralizedIntersectionUnion2019} proposed the GIoU term $\left( 1 - {GIoU} \left( {{\boldsymbol{l}_{x,y,m}},{{\boldsymbol{\hat l}}_{x,y,m}}}  \right)\right)$ to replace the term of $\sum\limits_{k = 1}^4 {{{\left( {l_{x,y,m}^{\left( k \right)} - \hat l_{x,y,m}^{\!\left( k \right)}} \!\right)} ^2}}$, where the GIoU calculation can be found in Appendix B. We adopt this idea. Therefore, the loss function of OBB regression at location ${\left( {x,y} \right)_m}$ in Eq.~\ref{eq:10} is obtained.

\textbf{2) PDF of object classification.} The object classification task in this case is composed of multiple i.i.d. binary classifications and each component of $\boldsymbol{y}$ is either 0 or 1. To estimate $\boldsymbol{y}$, the non-linear activation function $Sigmoid\left(  \cdot  \right)$ is used on the basic neuron model in output layers. Thus, each component of $\boldsymbol{\hat y}$ is in $\left( {0,1} \right)$ that represents the classification score. In CNN, this classification score is usually interpreted as “probability” of the binary classification \cite{renFasterRCNNRealTime2017a,redmonYouOnlyLook2016a}. Assume that, given $\boldsymbol{x}$ and $\boldsymbol{\theta}$, $\boldsymbol{y}$ follows $Bernoulli\left( {1,\boldsymbol{\hat y}} \right)$, and the PDF is
\begin{equation}
	\begin{array}{l}
		p\left( {\boldsymbol{y}\left| \boldsymbol{x} \right.;\boldsymbol{\theta} } \right) = {\boldsymbol{\hat y}}^{\boldsymbol{y}}{\left( {1 - \boldsymbol{\hat y}} \right)^{1 - \boldsymbol{y}}}.
	\end{array}
	\tag{A-8}
	\label{eq:A-8}
\end{equation}
Then, for multiple ${y^{\left( i \right)}},{\rm{ }}i = 1,2, \cdots ,m$, in different locations of output layers, their joint PDF is 
\begin{equation}
	\begin{array}{l}
	p\left( {{y^{\left( 1 \right)}} \cdots {y^{\left( m \right)}}\left| {{x^{\left( 1 \right)}} \cdots {x^{\left( m \right)}}} \right.;\theta } \right)\\
	= \prod\limits_{i = 1}^m {p\left( {{y^{\left( i \right)}}\left| {{x^{\left( i \right)}}} \right.;\theta } \right)}  \\
	= \prod\limits_{i = 1}^m {{{\left( {{{\hat y}^{\left( i \right)}}} \right)}^{{y^{\left( i \right)}}}}{{\left( {1 - {{\hat y}^{\left( i \right)}}} \right)}^{1 - {y^{\left( i \right)}}}}.} 
	\end{array}
	\tag{A-9}
	\label{eq:A-9}
\end{equation}
Thus, when $\boldsymbol{x}_{x,y,m}^{cls}$, $\boldsymbol{\theta}_{x,y,m}^{cls}$, and $\boldsymbol{obb}{_{x,y,m}}$ are given, the PDF of object classification is
\begin{equation}
	\begin{array}{l}
	\!\! p\left( {\boldsymbol{cls}{_{x,y,m}}\left| {\boldsymbol{obb}{_{x,y,m}};obj{_{x,y,m}};\boldsymbol{x}_{x,y,m}^{cls}} \right.;\boldsymbol{\theta} _{x,y,m}^{obb}} \right)\\
	\!\! =\! p(cls_{x,y,m}^{\left( 1 \right)} \cdots cls_{x,y,m}^{\left( {nu{m_{cls}}} \right)}\left| {\boldsymbol{obb}{_{x,y,m}};ob{j_{x,y,m}};} \right.\\
	\!\! x_{x,y,m}^{\left( 1 \right)}, \!\cdots x_{x,y,m}^{\left( {nu{m_{cls}}} \right)};{\theta {_{x,y,m}^{cls}}^{\left( 1 \right)}}, \!\cdots ,{\theta {_{x,y,m}^{cls}}\!^{\left( {nu{m_{cls}}} \right)}})\\
	\!\! = \!\prod\limits_{c = 1}^{\! num{_{\! cls}}} \!{{{\left( \!{\widehat {cls}_{x,y,m}^{\left( c \right)}} \right)}^{\! cls \!_{x,y,m}^{\left( c \right)}}} \! \times {{\left( {\!1 \!\! - \widehat {cls}_{x,y,m}^{\left( c \right)}} \! \right)}^{\!1 \!- \! {\! cls \!_{x,y,m}^{\left( c \right)}}}}}.
	\end{array}
	\tag{A-10}
	\label{eq:A-10}
\end{equation}

Similarly, when $x_{x,y,m}^{obj}$ and $\theta _{x,y,m}^{obj}$ are given. The PDF of $obj{_{x,y,m}}$ is
\begin{equation}
	\begin{array}{l}
	p\left( {ob{j_{x,y,m}}\left| {x_{x,y,m}^{obj}} \right.;\theta _{x,y,m}^{obj}} \right) = \\
	{\left( {{{\widehat {obj}}_{x,y,m}}} \right)^{ob{j_{x,y,m}}}} \times {\left( {1 - {{\widehat {obj}}_{x,y,m}}} \right)^{1 - ob{j_{x,y,m}}}},
	\end{array}
	\tag{A-11}
	\label{eq:A-11}
\end{equation}
where $\theta _{x,y,m}^{obj},{\rm{ }}m = 1,2,3,$ represent the parameter at ${\left( {x,y} \right)_m}$ used to predict whether this location is positive or negative.

\textbf{3) The joint PDF of the positive or negative location detection, OBB regression, and object classification.} Combining Eq.~\ref{eq:A-5}, Eq.~\ref{eq:A-10} and Eq.~\ref{eq:A-11}, we obtain Eq.~\ref{eq:18}.

\normalsize
\subsection{The calculation of GIoU in ORC}
The GIoU of Eq.~\ref{eq:11} in ORC is calculated according to Algorithm~\ref{alg:2}. 

\begin{algorithm}[htbp]
	\label{alg:2}
	\small%\footnotesize
	\caption{The calculation of GIoU in ORC}	
	\LinesNumbered
	\KwIn{The ground truth distances ${\boldsymbol{l}_{x,y,m}}$ composed of ${l_1}$, ${l_2}$, ${l_3}$, ${l_4}$, and the predicted distances ${\boldsymbol{\hat l}_{x,y,m}}$ composed of ${\hat l_1}$, ${\hat l_2}$, ${\hat l_3}$, ${\hat l_4}$.}
	\KwOut{$GIoU{_{x,y,m}}\left( {{\boldsymbol{l}_{x,y,m}},{{\boldsymbol{\hat l}}_{x,y,m}}} \right)$.}
	Area of ground truth HBB $area{_{x,y,m}} = \left( {{l_1} + {l_3}} \right) \times \left( {{l_2} + {l_4}} \right)$;\\
	Area of predicted HBB 
	${\widehat {area}_{x,y,m}} = \left( {{{\hat l}_1} + \hat l} \right) \times \left( {{{\hat l}_2} + {{\hat l}_4}} \right)$;\\
	Overlapping area 
	$area_{x,y,m}^{overlap} = \left( {\min \left( {{l_1},{{\hat l}_1}} \right) + \min \left( {{l_3},{{\hat l}_3}} \right)} \right)	\times \left( {\min \left( {{l_2},{{\hat l}_2}} \right) + \min \left( {{l_4},{{\hat l}_4}} \right)} \right)$;\\
	Area of the circumscribed HBB of the two HBBs above
	$area_{x,y,m}^{circ} = \left( {\max \left( {{l_1},{{\hat l}_1}} \right) + \max \left( {{l_3},{{\hat l}_3}} \right)} \right) \times \left( {\max \left( {{l_2},{{\hat l}_2}} \right) + \max \left( {{l_4},{{\hat l}_4}} \right)} \right)$;\\
	Area of the union region of the two HBBs above
	${U_{x,y,m}} = are{a_{x,y,m}} + {\widehat {area}_{x,y,m}} - area_{x,y,m}^{overlap}$;\\
	$IoU{_{x,y,m}}\left( {{\boldsymbol{l}_{x,y,m}},{\boldsymbol{\hat l}_{x,y,m}}} \right)
	=  \displaystyle{\frac{{area_{x,y,m}^{overlap}}}{U_{x,y,m}}}$;\\
	$GIoU{_{x,y,m}}\left( {{\boldsymbol{l}_{x,y,m}},{{\boldsymbol{\hat l}}_{x,y,m}}} \right)
	\! = IoU{_{x,y,m}}\left( {{\boldsymbol{l}_{x,y,m}},{{\boldsymbol{\hat l}}_{x,y,m}}} \right) \!- \frac{{area_{x,y,m}^{circ} \! - {U_{x,y,m}}}}{{area_{x,y,m}^{circ}}}$.
\end{algorithm}

%\bibColoredItems{blue}{wang2020learning, yang2021dense, yang2021learning, li2020object, xie2021oriented, cheng2021anchor, ma2021iqdet, ding2021object}

%\section*{Acknowledgement}
%\input{Acknowledgement}

%\section*{References}
\footnotesize%small%
\bibliographystyle{IEEEtran}
\bibliography{GGHL.bib}

%\input{Biography}
%\newpage

\end{document}